%% file: 2017-tbd-popovic.tex
\documentclass[twocolumn,natbib]{svjour3}

\makeatletter
\def\cl@chapter{\@elt {theorem}}
\makeatother

\usepackage{apalike}

\usepackage[printonlyused,withpage,nolist,nohyperlinks]{acronym}
\input{style/acronyms.tex}

\usepackage{textcomp}
\usepackage{bm}
\usepackage{gensymb}
\usepackage{graphicx}
\usepackage{amsmath}
\usepackage{algorithm}
\usepackage[noend]{algpseudocode}
\usepackage{subcaption}
\usepackage[rightcaption]{sidecap} \sidecaptionvpos{figure}{c}
\usepackage{wrapfig}
\usepackage{amsfonts}
\usepackage{siunitx}
\usepackage{booktabs}
\usepackage[font=small]{caption}
\usepackage[export]{adjustbox}
\usepackage{upgreek}
\usepackage{enumitem}
\usepackage{dblfloatfix}
\usepackage{tikz,tikz-3dplot}

\DeclareMathOperator*{\argmax}{\arg\!\max}

\DeclareMathOperator{\Tr}{Tr}

\usepackage{hyperref}
\hypersetup{colorlinks,breaklinks,
linkcolor=[rgb]{0.5,0,0},
citecolor=[rgb]{0.000,0.427,0.173},
urlcolor=[rgb]{0.031,0.318,0.612}}

\newcommand\edit[1]{{\color{black}#1}}
\newcommand\editrev[1]{{\color{black}#1}}

\usepackage[nameinlink]{cleveref}

\begin{document}
\title{An informative path planning framework \\ for UAV-based terrain monitoring}
\author{Marija Popovi\'{c}        \and
        Teresa Vidal-Calleja        \and
        Gregory Hitz                \and
        Jen Jen Chung               \and
        Inkyu Sa                    \and
        Roland Siegwart             \and
        Juan Nieto
        \thanks{* These authors were with Autonomous Systems  Lab., ETH Z\"{u}rich, Z\"{u}rich, Switzerland at the time of this work.}
}

\institute{Marija Popovi\'{c}*  \\
            \email{m.popovic@imperial.ac.uk}
             \at Smart Robotics  Lab.,
              Imperial College London,
              London, United Kingdom
             \and
           Teresa Vidal-Calleja
             \email{teresa.vidalcalleja@uts.edu.au}
             \at Centre for Autonomous Systems,
              University of Technology,
              Sydney, Australia
             \and
           Gregory Hitz*
           \at
              \email{gregory.hitz@sevensense.ch}
             \and
           Jen Jen Chung
           \at
              \email{jenjen.chung@mavt.ethz.ch}
             \and
           Inkyu Sa*
           \at
              \email{inkyu.sa@csiro.au}
               \at Dynamic Platforms, Robotics and Autonomous Systems Group,
               Data61, CSIRO, Brisbane, Australia
             \and
           Roland Siegwart
           \at
              \email{rsiegwart@ethz.ch}
             \and
           Juan Nieto \\
             \email{nietoj@ethz.ch}
             \at Autonomous Systems  Lab.,
              ETH Z\"{u}rich,
              Z\"{u}rich, Switzerland}

\date{Received: date / Accepted: date}

\maketitle

\begin{abstract}
Unmanned Aerial Vehicles (UAVs) represent a new frontier in a wide range of monitoring and research applications.
To fully leverage their potential,
a key challenge is planning missions for efficient data acquisition in complex environments.
To address this issue,
this article introduces a general Informative Path Planning (IPP) framework for monitoring scenarios using an aerial robot,
\edit{focusing on problems
in which the value of sensor information is unevenly distributed in a target area \editrev{and unknown \textit{a priori}}.}
The approach is capable of
\editrev{learning and focusing on regions of interest via adaptation to map}
either discrete or continuous variables on \edit{the} terrain
using variable-resolution data received from probabilistic sensors.
During a mission,
the terrain maps built online are used to plan information-rich trajectories in continuous 3-D space
by optimizing initial solutions obtained by a coarse grid search.
Extensive simulations show that our approach is more efficient than existing methods.
We also demonstrate its real-time application on a photorealistic mapping scenario
using a publicly available dataset
\edit{and demonstrate a proof of concept for an agricultural monitoring task}.

\end{abstract}

\section{Introduction} \label{S:introduction}

Autonomous mobile robots are increasingly employed to gather valuable scientific data about the Earth.
In the past several decades,
rapid technological advances have unlocked their potential as a flexible, cost-efficient tool
enabling monitoring at unprecedented levels of resolution and autonomy.
In many emerging aerial~\citep{Ezequiel2014,Vivaldini2016,Colomina2014} and aquatic~\citep{Hitz2014,Hitz2015,Girdhar2015} applications,
these devices are replacing traditional data acquisition campaigns
based on static sensors, manual sampling, or conventional manned platforms,
which can be unreliable, costly, and even dangerous~\citep{Dunbabin2012,Manfreda2018}.

The era of robotics-based monitoring has opened many interesting areas of research.
\edit{A key challenge arises in that practical devices are subject
to a finite quantity of sensing resources,
such as energy, time, or travel distance,
which limits the number of measurements that can be collected.
Therefore,
paths need to be planned to maximize the information gathered about an unknown environment
while satisfying the given budget constraint.}
This is known as the \emph{\ac{IPP}} problem,
which is the subject of much recent work.
\edit{Despite recent advances in this field,
most real robot systems today still perform data acquisition in a passive manner,
e.g., using coverage-based planning~\citep{Galceran2013},
as current \ac{IPP} solutions tend to be limited to
specific platforms or application domains.}

\edit{To address this,
our work introduces a general framework}
for surveying terrain characteristics using an aerial robot.
\edit{We present an \ac{IPP} strategy suitable for \emph{adaptive} scenarios,
which require modifying plans online
based on new measurements to \editrev{learn and} focus on targeted objects or regions of interest,
e.g., to find cars in a parking lot or map high-temperature areas.
This setup involves several closely related research challenges.}
First,
the dense visual information received from different altitudes
needs to be integrated into a compact probabilistic map.
Second,
using the map,
the planning unit must search for informative trajectories
in the large 3-D space above the monitored area,
which poses a complex optimization problem.
During this procedure,
a crucial aspect is trading off between sensor resolution and \ac{FoV},
while accounting for \edit{the platform-specific constraints and adaptivity requirements.
Finally,
the integrated system is required 
to be efficient for online mapping and planning
in a wide variety of scenarios.
In this paper,
we cater for these needs
by developing a modular \ac{IPP} framework
for aerial robots in general environmental monitoring applications.}

\subsection{Contributions} \label{ss:contributions}
Overall, this article corresponds to a major extension 
of \edit{the methods introduced} in the authors' preliminary \edit{conference} works.
\edit{The journal version brings together our previous contributions in
online informative planning~\citep{Popovic2017ICRA}
and multiresolution mapping~\citep{Popovic2017IROS}
with additional explanations and simulations
to consolidate the integrated system
along with open-source software.}

\edit{Building upon our developments in planning and mapping,
we present an \ac{IPP} framework} capable of mapping either discrete or continuous target variables on a terrain
using a probabilistic altitude-dependent sensor.
\edit{A key feature of the approach is that it is modular,
with interfaces and computational requirements that are easily adapted
for a given monitoring scenario.}
During a mission,
the planner uses the terrain maps built online to generate trajectories
in continuous 3-D space for maximized information gain.
\edit{This is achieved in a computationally efficient manner
by exploiting recursive Bayesian mapping methods
and an optimization strategy for \ac{IPP} with an informed initialization procedure.}
Our method was tested extensively in simulation
and \edit{its online and integration capabilities}
were validated by mapping a publicly available dataset
using an \ac{UAV} equipped with an image-based classifier.
\edit{Additionally, a proof of concept field deployment is presented
in an agricultural monitoring application
with real-time requirements.}

The core contributions of this work are:
\edit{\begin{enumerate}

 \item An informative planning framework that is applicable for mapping either discrete or continuous variables on a given terrain
 using probabilistic sensors for data acquisition in online settings
 with adaptivity requirements \editrev{by learning targeted objects or regions of interest}.


 \item The extensive evaluation of our framework in simulation,
 along with results from a publicly available dataset and field experiments to demonstrate its performance.

 \item The release of our implementation as an open-source software package\footnote{Available at: \url{github.com/ethz-asl/tmplanner}.} for usage and further development by the community.

\end{enumerate}}

The remainder of this paper is organized as follows.
\Cref{S:related_work} discusses prior studies relevant to our work.
We define the \ac{IPP} problem in \Cref{S:problem_statement}
and detail our proposed methods for mapping and planning in \Cref{S:mapping_approach,S:planning_approach}, respectively.
Our experimental results are reported \editrev{in \Cref{S:results,S:field_deployments}}.
Finally, in \Cref{S:conclusion},
we conclude with an outlook towards future work.

\section{Related work} \label{S:related_work}
A large and growing body of literature exists for autonomous information gathering problems.
Recently,
this field has attracted considerable interest in the context of robotics-based environmental monitoring~\citep{Dunbabin2012}
for a wide variety of applications,
including aerial surveillance~\citep{Colomina2014,Vivaldini2016},
aquatic monitoring~\citep{Hitz2014,Hitz2015},
and infrastructure inspection~\citep{Ezequiel2014,Bircher2016a}.
This section overviews recent work based on two main research streams:
(1) methods for environmental modeling
and (2) algorithms for informative planning, or efficient data acquisition.

\subsection{Environment mapping}

In data gathering scenarios,
a model of the environment is fundamental to capture the target variable of interest.
Occupancy grids are the most commonly-used representation
for spatial sensing with uncorrelated measurements~\citep{Elfes1989}.
This type of model is suitable for active classification problems with discrete labels,
such as occupancy mapping~\citep{Charrow2015} and semantic segmentation~\citep{Berrio2017},
and offers relatively high computational efficiency.

However, many natural phenomena
exhibit complex interdependencies
where the assumption of independent measurements does not hold.
A popular Bayesian technique for handling such relationships is using \acp{GP}~\citep{Rasmussen2006}.
For \ac{IPP}, they have been applied in various scenarios~\citep{Hollinger2014,Hitz2015,Binney2012,Wei2016}
to collect data accounting for map structure and uncertainty.
This framework permits using different kernel functions
to express data relations within the environment~\citep{Singh2010}
and approximations to scale to large datasets~\citep{Rasmussen2006}.

\edit{For terrain monitoring,
the main issue with applying \acp{GP} directly is
the escalating computational load as dense imagery data accumulate over time.
Moreover,
limited studies have addressed data fusion using images at varying resolutions and noise levels.
In our prior work,
we addressed this
by introducing a multiresolution \ac{GP}-based mapping approach for \ac{IPP}
based on a Bayesian filtering technique, inspired by~\citet{Vidal-Calleja2014}.
This method replaces the complexity of conventional \ac{GP} regression
with constant processing time in the number of measurements
and cubic scaling only with the size of the fixed terrain map.}
Our current work extends these ideas
by presenting a general framework that can handle both non-correlated and correlated monitored variables
\edit{and by demonstrating its integration with a practical altitude-dependent sensor.}

\subsection{Informative planning}

We also distinguish between (i) non-adaptive and (ii) adaptive planning strategies.
Non-adaptive approaches, e.g., coverage methods~\citep{Galceran2013}, explore an environment
using a sequence of pre-determined actions.
Adaptive approaches~\citep{Hitz2015,Girdhar2015,Lim2015,Hollinger2013}
allow plans to change as information is collected,
\edit{making them suitable for planning based on targeted application-specific interests}.

In its most general form,
the data gathering task amounts to one of sequential decision-making under uncertainty,
which can be expressed as a \ac{POMDP}~\citep{Kaelbling1998}.
Unfortunately,
despite substantial progress in recent years~\citep{Chen2016,Kurniawati2008},
solving large-scale \ac{POMDP} models remains an open challenge.
\edit{For data gathering,
a major issue with practical algorithms is that
they do not generalize well for large state spaces or over long horizons~\citep{Lim2015a}.
Although approximate \ac{MDP}-based solvers have been successfully applied for online planning,
their performance is limited in terrain monitoring setups
since they cannot exploit the ability of \acp{UAV} to gather multiresolution information from different altitudes,
e.g., to adaptively find objects of interest \citep{Arora2018}.
The computational and representational issues associated with such methods
motivates more efficient \ac{IPP}-based solutions.}

The NP-hard sensor placement problem
addresses selecting the most informative measurement sites in a static setting
\edit{subject to cardinality constraints~\citep{Krause2008}.
The related \ac{IPP} problem builds upon this task
by maximizing the information gathered
\editrev{by a mobile platform traversing a path while subject to finite resources}.
A number of general algorithms for \ac{IPP} operate by
performing combinatorial optimization over a discrete grid.
The seminal work of~\citet{Chekuri2005} presents
a recursive greedy algorithm providing a sub-logarithmic approximation in sub-exponential time.
\citet{Singh2009} apply these ideas to environmental sensing problems
and extend them to multi-robot settings.
More recently, 
branch and bound techniques have been proposed to improve computational efficiency 
using monotonic objective functions~\citep{Binney2012}.
In 2-D scenarios,
an alternative strategy is to represent the robot workspace
as decomposed cells~\citep{Cai2009} or a connectivity tree~\citep{Ferrari2009}.
These methods~\citep{Cai2009,Binney2012} have been shown to outperform
\editrev{coverage-based planning strategies~\citep{Choset2001,Galceran2013}}
and are applicable in cluttered environments.

However,
discrete solvers are typically limited in resolution
and scale exponentially with the problem instance.
To reduce the computational overhead,
we follow previous approaches~\citep{Bircher2016a,Hitz2014}
in using a method with a limited look-ahead
to generate plans within a fixed horizon of the total budget.
This approach allows us to perform incremental replanning,
as required for adaptive settings,
by trading off guarantees on exploration optimality outside the planning horizon.}

Continuous-space planning strategies
offer better scalability
\edit{in comparison to discrete methods}
by leveraging \edit{\ac{RRT}-like} sampling-based methods~\citep{Hollinger2014}
or splines~\citep{Vivaldini2016,Hitz2015,Charrow2015,Morere2017} directly in the robot workspace.
As in our prior work~\citep{Popovic2017ICRA},
we follow the latter approaches in defining smooth polynomial trajectories~\citep{Richter2013}
which are optimized globally for an information objective.
Our spline optimization problem setup most closely resembles the ones studied by
~\citet{Hitz2015} and~\citet{Morere2017};
however,
our strategy differs in that it uses an informed initialization procedure
to obtain faster convergence.

\edit{Specifically,
we follow a two-step approach,
which exploits a greedy, grid-based solution
to the resource-constrained \ac{IPP} problem
to initialize an evolutionary optimization routine.
Our grid search procedure closely resembles frontier-based strategies
commonly used for exploration in cluttered environments~\citep{Yamauchi1998}.
However, a key difference is that our method does not rely on
the internal geometry of the environment
as the camera \ac{FoV} is unobstructed by obstacles.
For the optimization step,
an alternative approach to our global evolutionary strategy
is to generate control actions for the robot
using gradient ascent on an information objective~\citep{Schwager2017,Julian2011,Grocholsky2003}.
Such techniques would be useful when the initial grid-based solution
is close to a trajectory with high information quality,
but may be susceptible to local minima.
This issue can be addressed by applying
algorithms based on random walks, local roadmaps~\citep{Lu2014,Bellini2014},
or two-stage optimization strategies~\citep{Charrow2015,Rastogoftar2018}.}

\edit{
The motivation behind our approach is shared by the studies of~\citet{Jadidi2016,Jadidi2018} in robotic exploration and information gathering.
In contrast, however,
we address the problem of terrain monitoring, considering a 3-D robot workspace, 
instead of 2-D occupancy mapping and environmental mapping with a point-based sensor.

Within this context,
several recent works have examined applications
similar to ours. Notably, }
\citet{Sadat2015} devise an adaptive planner
\edit{for aerial coverage problems in which regions of interest are non-uniformly distributed.}
Their method, however, assumes discrete viewpoints
and does not support probabilistic data acquisition.
\edit{To address this,
\citet{Bellini2014} present an information gradient-based control law
for active target classification in cluttered environments.
More similarly to our approach,
\citet{Arora2017} introduce an efficient near-optimal algorithm
that provides anytime solutions in adaptive scenarios.
As further discussed by~\citet{Arora2018},
their setup considers using a multiresolution sensor
to gather targeted information about specific objects
as they are detected.
A key difference
is that our method also caters for probabilistic variations in sensor \emph{noise} for data fusion at different altitudes,
and can accommodate adaptively mapping continuous variables, e.g., temperature,
as well as discrete objects, e.g., cars.}

\section{Problem statement} \label{S:problem_statement}

Our setup focuses on efficient data-gathering strategies
for an aerial robot
operating above a terrain. 
The aim is to maximize the information collected about the environment,
while respecting resource constraints,
such as energy, time, or distance budgets.
Formally, this is known as the \ac{IPP} problem,
which is defined as follows.
We seek an optimal trajectory $\psi^*$
in the space of all continuous trajectories $\Psi$
for maximum gain in some information-theoretic measure:
\begin{equation}
\begin{aligned}
  \psi^* ={}& \underset{\psi \in \Psi}{\argmax} \,
\frac{\mathrm{I}(\textsc{measure}(\psi))}{\textsc{cost}(\psi)} \, \textrm{,} \\
 & \text{s.t. } \textsc{cost}(\psi) \leq B \, \textrm{.}
 \label{E:ipp_problem}
\end{aligned}
\end{equation}
The function \textsc{measure(\textperiodcentered)} obtains a finite set of measurements along
trajectory $\psi$ in the 3-D space above the environment,
and \textsc{cost(\textperiodcentered)} provides the corresponding
cost,
which cannot exceed a predefined budget $B$.
The operator $\mathrm{I}($\textperiodcentered$)$ defines the information objective
quantifying the utility of the acquired measurements.

\section{Mapping approach} \label{S:mapping_approach}
\begin{figure*}[!h]
\centering
\includegraphics[width=0.9\textwidth]{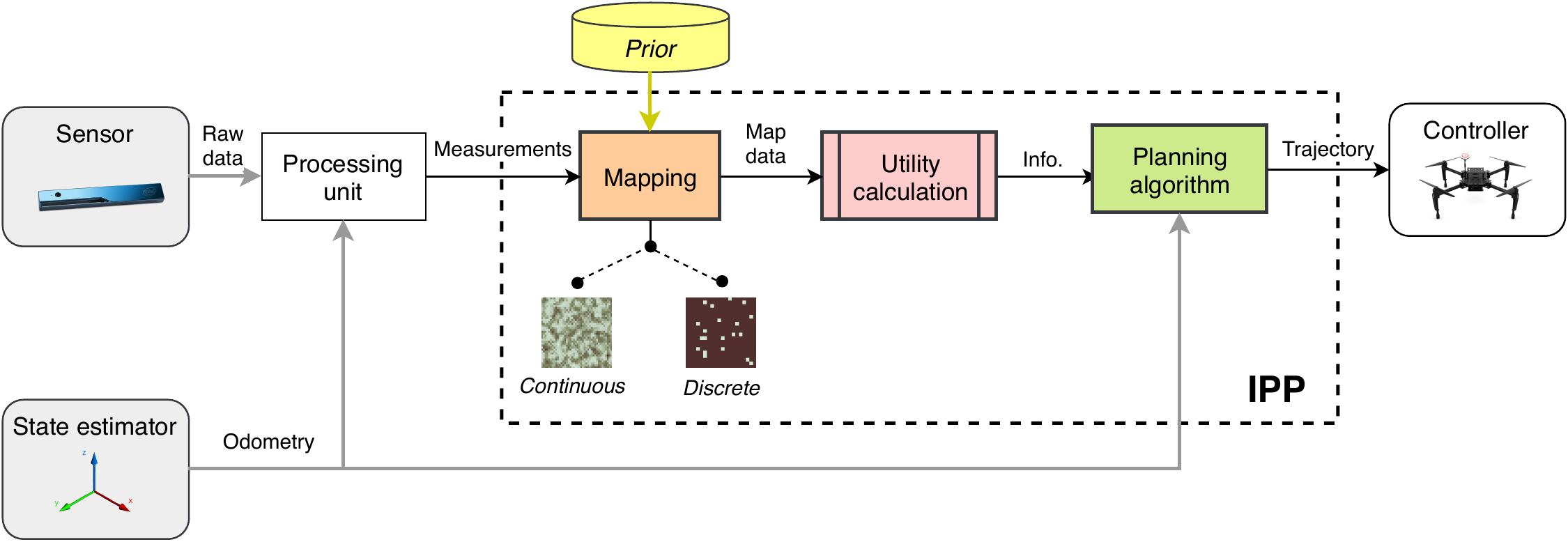}
 \caption{System diagram showing the key elements of our IPP framework.
 A map of the target environment is built using measurements extracted
 from a sensor data stream.
 \editrev{The dashed lines indicate that the map representation can be either discrete or continuous.}
 At a particular time instant,
 the map knowledge is used by the planning unit
 to find the most useful trajectories
 for data collection,
 starting at the current pose.
 These are then executed by the robot,
 allowing for subsequent map updates
 in a closed-loop manner.
 } \label{F:ipp_system}
\end{figure*}

In this section, we propose
a new mapping framework for terrain monitoring applications.
The generic structure of our system setup
is depicted in \Cref{F:ipp_system}.
As shown,
our framework is capable of mapping either discrete or continuous variables
based on measurements extracted from a sensing unit,
e.g., a depth or multispectral camera.
For a particular problem setup,
the map representation can be selected
depending on the type of data received.
During a mission, the planner uses the terrain maps built online
to optimize continuous trajectories
for maximum gain in an information metric
reflecting the mission aim.
A key aspect of our architecture is its generic formulation,
which enables it to adapt to any surface mapping scenario,
e.g., elevation~\citep{Colomina2014}, pipe thickness~\citep{Vidal-Calleja2014},
gas concentration~\citep{Marchant2014},
spatial occupancy~\citep{OCallaghan2012}, seismic hazards~\citep{Gao2017},
post-disaster assessment~\citep{Ezequiel2014}, signal strength~\citep{Hollinger2014},
etc.

In \Cref{SS:mapping_discrete,SS:mapping_continuous}, we present methods for map representation
for monitoring discrete and continuous targets, respectively,
as the basis of our framework.
In \Cref{S:planning_approach}, these concepts are used to formulate the objective function,
and we describe our adaptive planning scheme.

\subsection{Discrete variable mapping} \label{SS:mapping_discrete}
We study the task of monitoring a discrete variable
as an active classification problem.
The terrain environment $\mathcal{E}$
is discretized and represented using a 2-D occupancy map $\mathcal{X}$~\citep{Elfes1989},
where each grid cell is associated with an independent Bernoulli random variable
indicating the probability of target occupancy
(e.g., presence of weed on a farmland).
Measurements are taken with a downwards-looking sensor
providing inputs to a data processing unit,
from which discrete classification labels are obtained.
At time $t$,
for each observed cell $\mathbf{x}_i \in \mathcal{X}$ within the \ac{FoV}
from a \ac{UAV} pose $\mathbf{p}$ above the terrain,
a log likelihood update is performed given an observation $z$:
\begin{equation}
  L(\mathbf{x}_i\,|\,z_{1:t}, \mathbf{p}_{1:t}) =
L(\mathbf{x}_i\,|\,z_{1:t-1}, \mathbf{p}_{1:t-1}) +
L(\mathbf{x}_i\,|\,z_t, \mathbf{p}_t) \, \text{,} \label{E:occupancy_grid_update}
\end{equation}
where $L(\mathbf{x}_i\,|\,z_t, \mathbf{p}_t)$ denotes
the altitude-dependent inverse sensor model log-likelihood
capturing the classification output.

\begin{SCfigure}[][!h]
  \centering
  \includegraphics[width=0.5\columnwidth]{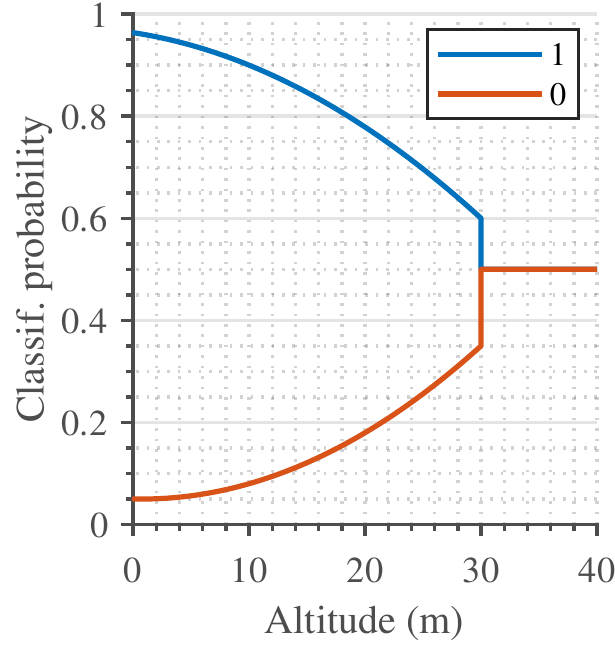}
   \caption{Sensor model
   for a typical camera-based binary classifier
   operating above a terrain.
   The blue and orange curves depict the probability of observing label `1'
   for a map cell containing `1' or `0', respectively,
   i.e., $p(z\,|\,\mathbf{x}_i, \mathbf{p})$.
   As altitude increases, the curves approach unknown classification
   probability (0.5).}\label{F:sensor_binary}
\end{SCfigure}

As an example,
\Cref{F:sensor_binary} shows the sensor model
for a hypothetical camera-based binary classifier
labeling observed cells
as `1' (occupied by target) or `0' (target-free).
For each class,
curves are defined to account for poorer classification
with lower-resolution measurements taken at higher altitudes.
In practice, these curves can be obtained
through a Monte Carlo-type accuracy analysis of raw classifier data
by averaging the number of true and false positives
(blue and orange curves, respectively)
recorded at different altitudes.

The described approach can be easily extended to mapping multiple class labels
by maintaining layers of occupancy maps for each,
as demonstrated in \Cref{SS:dataset_evaluation}.

\subsection{Continuous variable mapping} \label{SS:mapping_continuous}
To monitor a continuous variable,
our framework leverages a more sophisticated mapping method using \acp{GP}
to encode spatial correlations common in environmental distributions.
We use a \ac{GP} to initialize a recursive filtering procedure
with probabilistic sensors at potentially different resolutions.
This approach replaces the computational burden of applying \acp{GP} directly 
with constant processing time in the number of measurements.
We describe our method for creating prior maps
before detailing the Bayesian data fusion technique.

\subsubsection{Gaussian processes} \label{SSS:gps}
A \ac{GP} is used to model spatial correlations on the terrain
in a probabilistic and non-parametric manner~\citep{Rasmussen2006}.
The target variable for mapping
is assumed to be a continuous function in 2-D space:
$\zeta: \mathcal{E} \rightarrow \mathbb{R}$.
Using the \ac{GP}, a Gaussian correlated prior is placed over the function space,
which is fully characterized by the mean $\boldsymbol{\upmu} = \mathbb{E}[\boldsymbol{\upzeta}]$
and covariance $\mathbf{P} = \mathbb{E}[(\boldsymbol{\upzeta}-\boldsymbol{\upmu})(\boldsymbol{\upzeta}^\top-\boldsymbol{\upmu}^\top)]$
as $\boldsymbol{\upzeta} \sim \mathcal{GP}(\boldsymbol{\upmu},\,\mathbf{P})$,
where $\mathbb{E}[$\textperiodcentered$]$ denotes the expectation operator.

Given a pre-trained kernel $K(\mathcal{X},\mathcal{X})$ for a fixed-size terrain
discretized at a certain resolution with a set of $n$ locations $\mathcal{X} \subset \mathcal{E}$,
we first specify a finite set of new prediction points $\mathcal{X}^* \subset \mathcal{E}$
at which the prior map is to be inferred.
For unknown environments, as in our setup,
the values at $\mathbf{x}_i \in \mathcal{X}$ are initialized uniformly
with a constant prior mean.
For known environments,
the \ac{GP} can be trained from available data
and inferred at the same or different resolutions.
The covariance is calculated
using the classic \ac{GP} regression equation~\citep{Reece2013}:
\begin{equation}
\begin{split}
 \mathbf{P} ={}& K(\mathcal{X}^*,\mathcal{X}^*) - K(\mathcal{X}^*,\mathcal{X})[K(\mathcal{X},\mathcal{X}) +
 \sigma_n^2\mathbf{I}_n]^{-1} \\ 
 &\times K(\mathcal{X}^*,\mathcal{X})^\top \, \textrm{,}
\end{split} \label{E:gp_cov}
\end{equation}
where $\mathbf{P}$ is the posterior covariance,
$\mathbf{I}_n$ is the $n$\,$\times$\,$n$ identity matrix,
$\sigma_n^2$ is a hyperparameter representing signal noise variance,
and $K(\mathcal{X}^*,\mathcal{X})$ denotes cross-correlation terms
between the predicted and initial locations.

The kernel $K($\textperiodcentered$)$
determines the generalization properties of the \ac{GP} model,
and is chosen to describe the characteristics of $\zeta$.
To describe environmental phenomena,
we suggest choosing from among a number of well-known kernel functions
common in geostatistical analysis,
e.g., the squared exponential or Mat\'ern functions.
The free parameters of the kernel function,
called hyperparameters, control relations within the \ac{GP}.
These values can be optimized using various methods~\citep{Rasmussen2006}
to match the properties of $\zeta$ by training on multiple maps obtained previously
at the required resolution.

Once the correlated prior map $p(\boldsymbol{\upzeta}\,|\,\mathcal{X})$ is determined,
independent noisy measurements at variable resolutions
are fused as described in the following section.

\subsubsection{Sequential data fusion} \label{SSS:data_fusion}
A key component of our framework
is a map update procedure based on recursive filtering.
Given a uniform mean and the spatial correlations
captured by \Cref{E:gp_cov},
the map $p(\boldsymbol{\upzeta}\,|\,\mathcal{X}) \sim \mathcal{GP}(\boldsymbol{\upmu}^-,\,\mathbf{P}^-)$
is used as a prior for fusing new sensor measurements.

Let $\mathbf{z} = [z_1, \ldots, z_m]^\top$ denote new $m$ independent measurements
received at points $[\mathbf{x}_1,\ldots,\mathbf{x}_m]^\top \subset \mathcal{X}$
modeled assuming a Gaussian sensor
as $p(z_i\,|\,\zeta_i, \mathbf{x}_i) = \mathcal{N}(\mu_{s,i}, \sigma_{s,i})$.
To fuse the measurements $\mathbf{z}$
with the prior $p(\boldsymbol{\upzeta}\,|\,\mathcal{X})$,
we use the maximum \emph{a posteriori} estimator:
\begin{equation}
 \argmax_{\boldsymbol{\upzeta}} p(\boldsymbol{\upzeta}\,|\,\mathbf{z},\mathcal{X}) \, \textrm{.}
\end{equation}
The \ac{KF} update equations
are applied directly to compute the posterior density
$p(\boldsymbol{\upzeta}\,|\,\mathbf{z},\mathcal{X}) \propto p(\mathbf{z}\,|\,\boldsymbol{\upzeta},\mathcal{X}) \times
p(\boldsymbol{\upzeta}\,|\,\mathcal{X}) \sim \mathcal{GP}(\boldsymbol{\upmu}^+,\,\mathbf{P}^+)$~\citep{Reece2013}:
\begin{align}
 \boldsymbol{\upmu}^+ ={}& \boldsymbol{\upmu}^- + \mathbf{K}\mathbf{v} \, \label{E:kf_mean} \textrm{,} \\
 \mathbf{P}^+ ={}& \mathbf{P}^- - \mathbf{K}\mathbf{H}\mathbf{P}^- \, \textrm{,} \label{E:kf_cov}
\end{align}
where $\mathbf{K} = \mathbf{P}^-\mathbf{H}^\top \mathbf{S}^{-1}$ is the Kalman gain,
and $\mathbf{v} = \mathbf{z} - \mathbf{H}\boldsymbol{\upmu}^-$
and $\mathbf{S} = \mathbf{H}\mathbf{P}^-\mathbf{H}^\top + \mathbf{R}$
are the measurement and covariance innovations.
$\mathbf{R}$ is a diagonal $m \times m$ matrix of altitude-dependent variances $\sigma^2_{s,i}$
associated with each measurement $z_i$,
and $\mathbf{H}$ is an $m \times n$ matrix denoting a linear measurement model
that intrinsically selects part of the state $\{\zeta_1,\ldots,\zeta_m\}$ observed through $\mathbf{z}$.
The information to account for variable-resolution measurements
is incorporated according to the measurement model $\mathbf{H}$
in a simple manner as detailed in the following section.

The constant-time updates in \Cref{E:kf_mean,E:kf_cov}
are repeated every time new data are registered.
Note that, as all models are linear in this case,
the \ac{KF} update produces the optimal solution.
Moreover, this approach is agnostic to the type of sensor used
as it permits fusing heterogeneous data into a single grid map.

\subsubsection{Altitude-dependent sensor model} \label{SSS:sensor_continuous}
As an example,
we detail an altitude-dependent sensor model
for a downward-facing camera
used to take measurements of a terrain,
e.g., a farmland or disaster site.
In contrast with the pure classification case in \Cref{SS:mapping_discrete},
our model needs to express uncertainty
with respect to a continuous target distribution.
To do this,
we consider that the visual data
degrades with altitude in two ways:
(i)~noise and (ii)~resolution.
The proposed model accounts for these issues
in a probabilistic manner as follows.

We assume an altitude-dependent Gaussian sensor noise model.
For each observed point $\mathbf{x}_i \in \mathcal{X}$,
the camera provides a measurement $z_i$
capturing the target field $\zeta_i$ as
$\mathcal{N}(\mu_{s,i}, \sigma_{s,i})$, where $\sigma_{s,i}^2$
is the noise variance expressing uncertainty in $z_i$.
To account for lower-quality images
taken with larger camera footprints,
$\sigma_{s,i}^2$ is modeled as increasing with the \ac{UAV} altitude $h$ using:
\begin{equation}
 \sigma_{s,i}^2 = a(1-e^{-bh}) \, \textrm{,} \label{E:sensor_continuous}
\end{equation}
where $a$ and $b$ are positive constants.

\begin{SCfigure}[][!h]
 \centering
  \includegraphics[width=0.5\columnwidth]{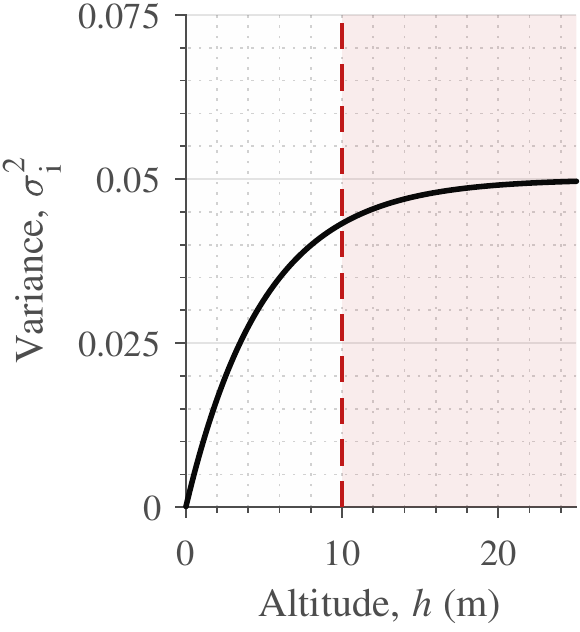}
   \caption{Inverse sensor noise model for a camera providing measurements as $\mathcal{N}(\mu_{s,i},
   \sigma_{s,i}^2)$ with $a = 0.2$, $b = 0.05$ in \Cref{E:sensor_continuous}.
   The uncertainty $\sigma_{s,i}^2$ increases with $h$ to represent degrading image quality.
   The dotted line at $h = 10$\,m indicates the altitude above which image resolution scales down by a factor of 2.
   }\label{F:sensor_continuous}
\end{SCfigure}

\Cref{F:sensor_continuous} illustrates the sensor noise model
used to evaluate our setup in \Cref{S:results}
which represents a camera.
The measurements $z_i$
denote the levels of the continuous variable being surveyed,
e.g., green biomass level or temperature.
\edit{This model is designed so that
the quality of the sensor data decreases
at higher altitudes,
according to requirements of the terrain monitoring problem.}
As for the discrete classifier in \Cref{SS:mapping_discrete},
\edit{in practice,
the parameters of such a model} can be obtained
\edit{by analyzing how the sensor behaves
at different altitudes
using previously acquired datasets}.

We define altitude envelopes corresponding to different resolution scales
with respect to the initial points $\mathcal{X}$ on the map.
This is motivated by the fact that the \ac{GSD} ratio (in m$/$px)
depends on the altitude of the sensor and its fixed intrinsic resolution.
To handle data received from variable altitudes,
adjacent $\mathbf{x}_i$ are indexed
by a single sensor measurement $z_i$
through the measurement model $\mathbf{H}$.
At lower altitudes
(higher \acp{GSD}, corresponding to the maximum mapping resolution in $\mathcal{X}$),
$\mathbf{H}$ is simply used
to select the part of the state observed
with a scale of $1$.
However, at higher altitudes (lower \acp{GSD}),
the elements of $\mathbf{H}$ used to map multiple $\zeta_i$ to a single $z_i$
are scaled by the square inverse of the resolution scaling factor $s_f$.
Note that the fusion procedure described in \Cref{SSS:data_fusion}
is always performed at the maximum mapping resolution \edit{as defined by $\mathcal{X}$},
so that the proposed model $\mathbf{H}$
considers low-resolution measurements
as a scaled average of the high-resolution map.

\edit{\Cref{F:mapping_approach} consolidates our strategy with an example.
The map in (d) depicts the ground truth
corresponding to a field distribution on a terrain.
Mapping is performed at the resolution shown in (a),
where the grid cells correspond to the locations in $\mathcal{X}$.
The maps in (e)-(h) demonstrate sequentially fusing
two measurements taken at different altitudes (middle row)
into a single probabilistic map (bottom row),
i.e., (g) and (h) visualize the results of fusing first (e) then (f), respectively,
assuming map initialization with a uniform mean.
The plots in (b) and (c) (top row) schematize the \ac{UAV} poses
from which the measurements were registered,
with the red grids indicating their resolutions.
In (c),
the scaling factor is $s_f = 0.5$,
such that 4 locations in $\mathcal{X}$ (black cells) map to a single measurement (red cell).
By inspecting the final field map in (h), upon fusing (f),
it can be seen the off-center values are more widely diffused compared to those in the center,
where the higher-quality measurement in (e) was taken,
as expected.

\begin{figure}[!h]

\centering

  \begin{subfigure}[]{0.3\columnwidth}
  \centering
  \vspace*{5mm}
  \input{additionals/gt_tikz.tex}
  \caption{}
  \end{subfigure} \hspace{0.3mm}
  \begin{subfigure}[]{0.3\columnwidth}
  \centering
  \vspace*{5mm}
  \input{additionals/low_altitude_tikz.tex}
  \caption{}
  \end{subfigure} \hspace{0.3mm}
  \begin{subfigure}[]{0.3\columnwidth}
  \centering
  \input{additionals/high_altitude_tikz.tex}
  \caption{}
  \end{subfigure}

\centering

  \vspace*{3mm}
  \begin{subfigure}[]{0.23\columnwidth}
  \includegraphics[width=\columnwidth]{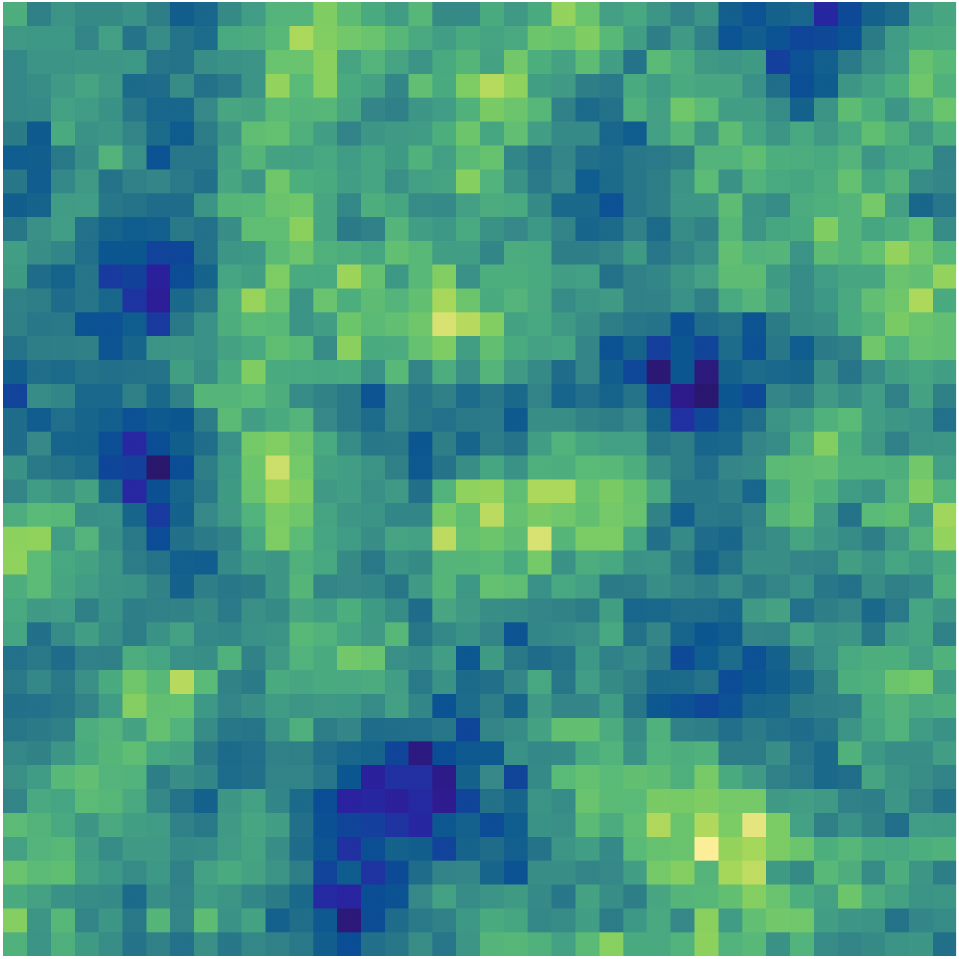}
  \caption{}
  \end{subfigure} \hspace{6mm}
  \begin{subfigure}[]{0.23\columnwidth}
  \includegraphics[width=\columnwidth]{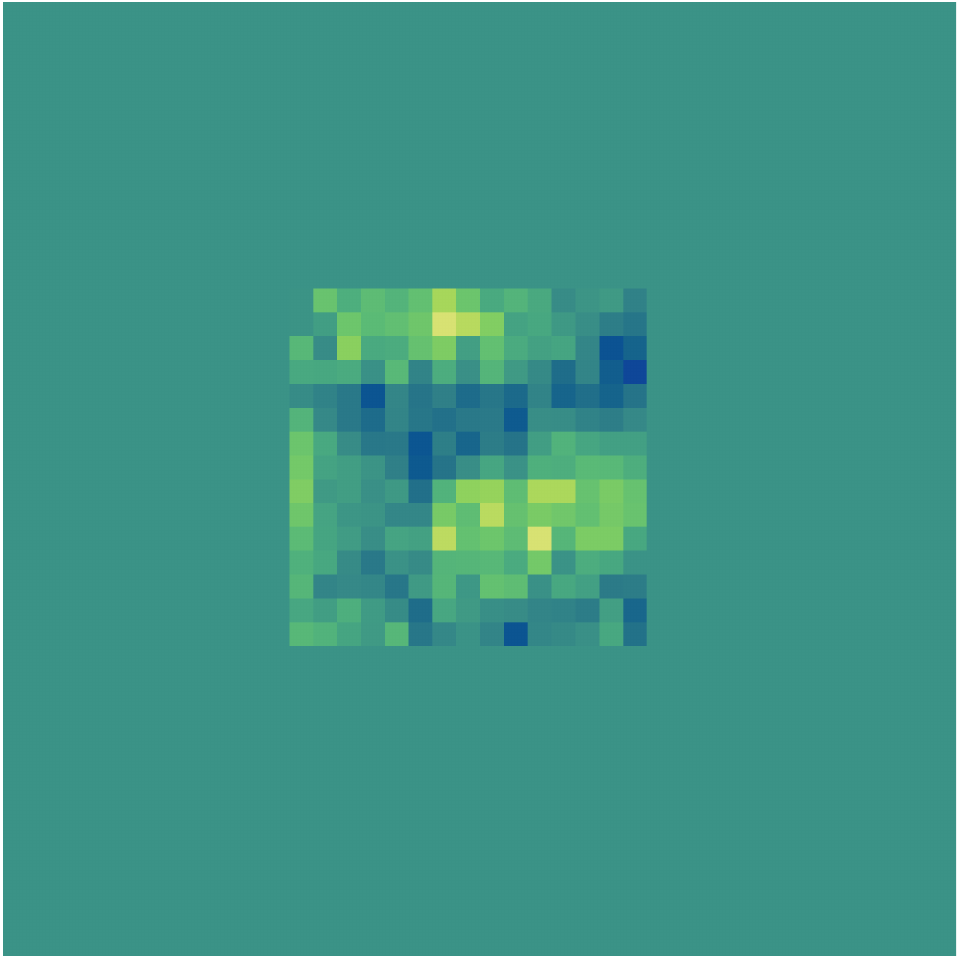}
  \caption{}
  \end{subfigure} \hspace{6mm}
  \begin{subfigure}[]{0.23\columnwidth}
  \includegraphics[width=\columnwidth]{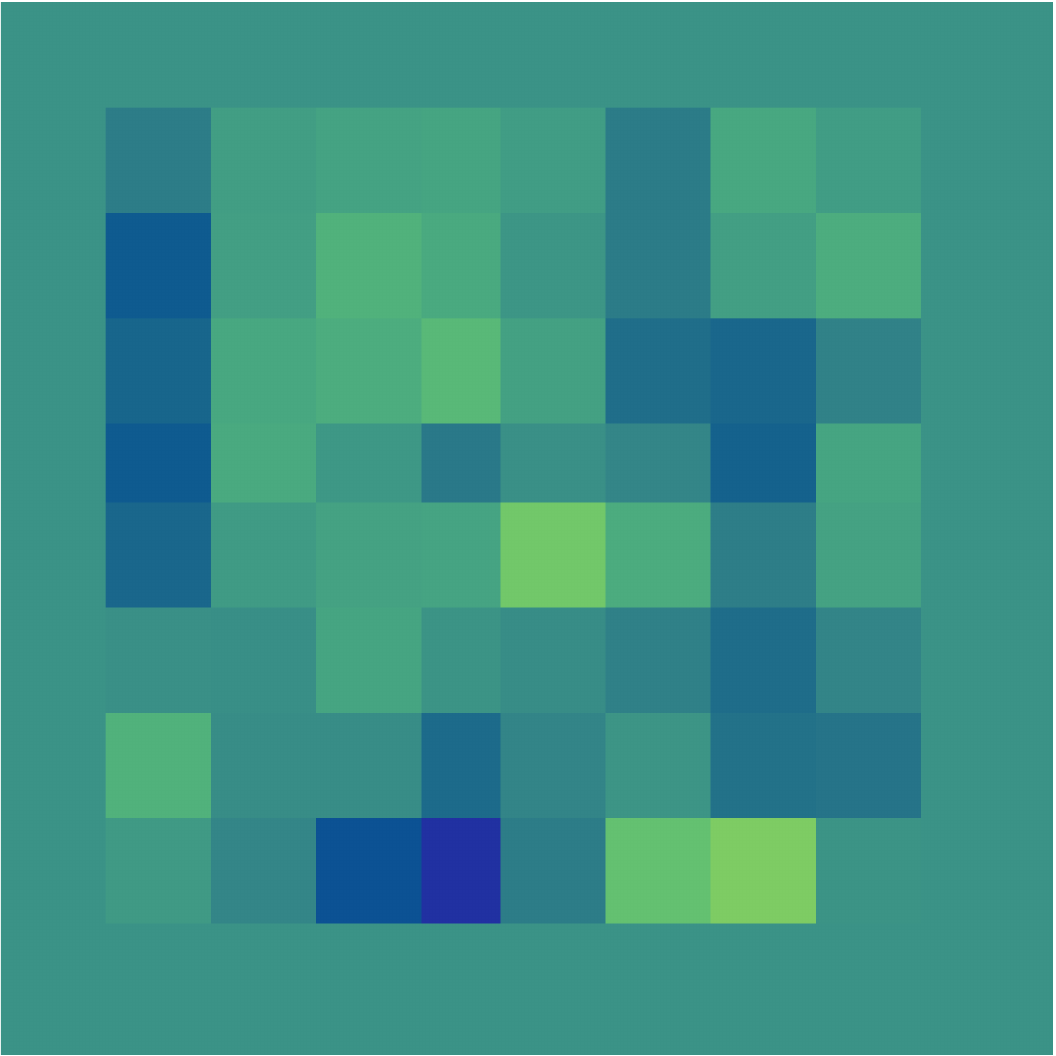}
  \caption{}
  \end{subfigure}
    
  \vspace*{3mm}
  \begin{subfigure}[]{0.23\columnwidth}
  \hspace{8mm}
  \includegraphics[height=\columnwidth]{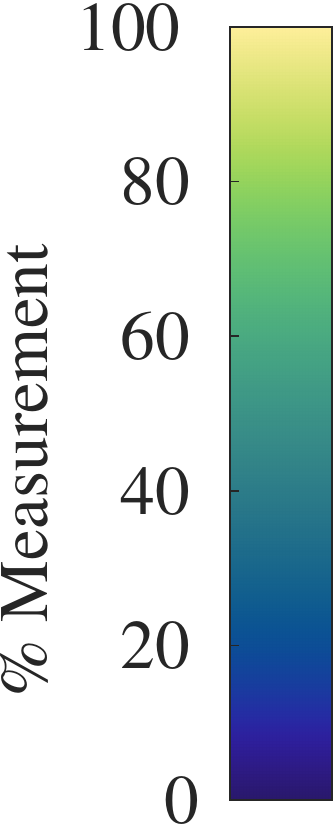}
  \vspace{5mm}
  \end{subfigure} \hspace{6mm}
  \begin{subfigure}[]{0.23\columnwidth}
  \includegraphics[width=\columnwidth]{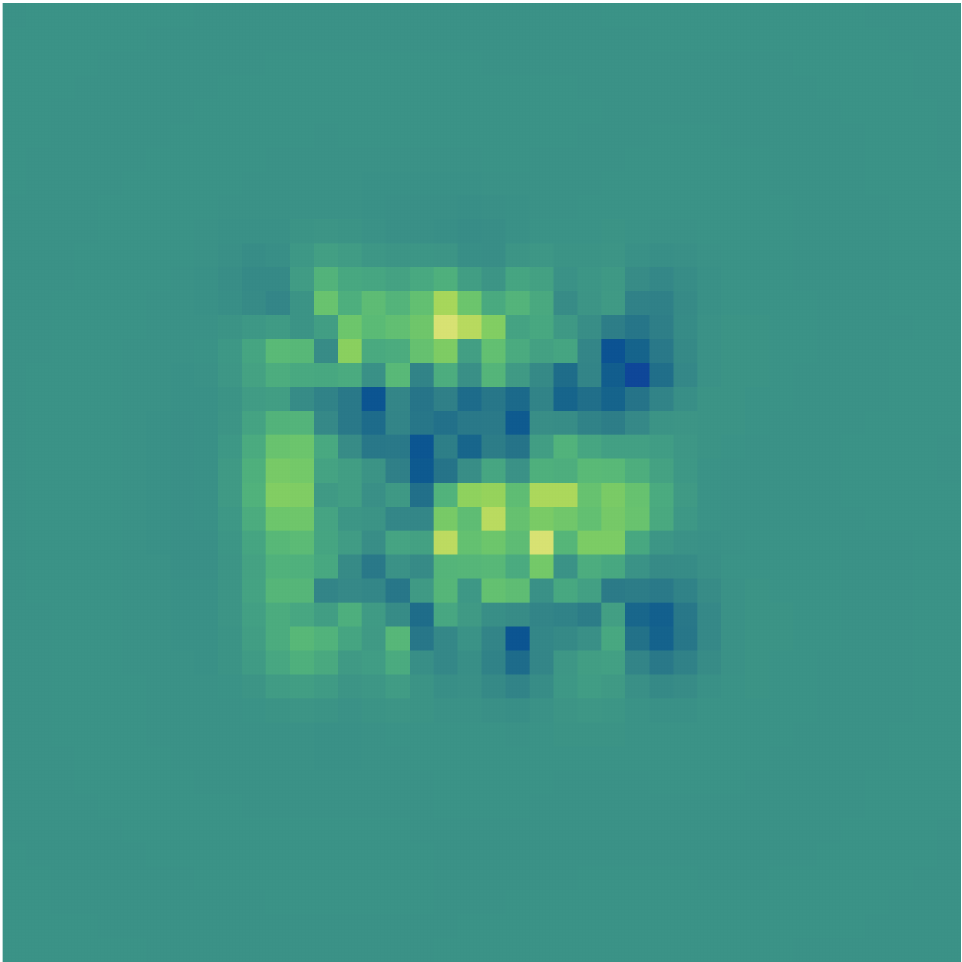}
  \caption{}
  \end{subfigure} \hspace{6mm}
  \begin{subfigure}[]{0.23\columnwidth}
  \includegraphics[width=\columnwidth]{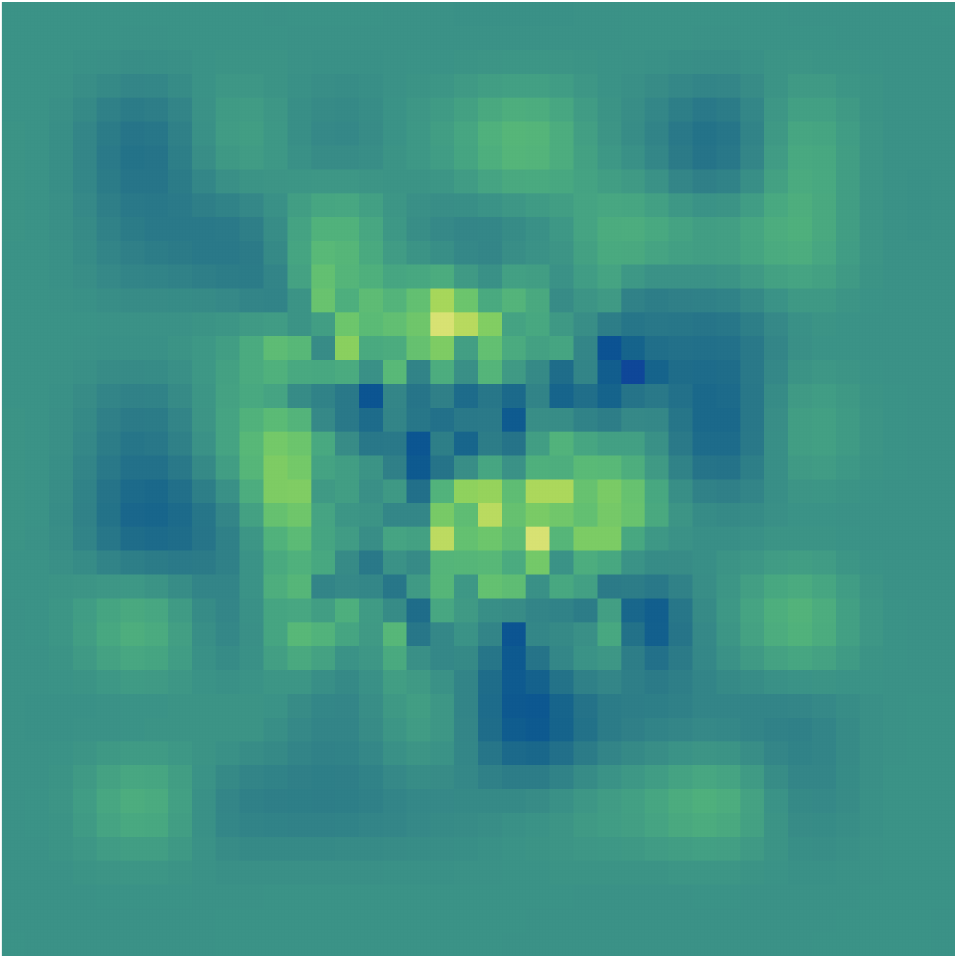}
  \caption{}
  \end{subfigure}
   \caption{Overview of our strategy for continuous variable mapping. A ground truth field is shown in (d). (e) and (f) depict measurements taken from $9$\,m and $20$\,m altitudes, with different resolutions visible in the projected camera \acp{FoV} (pyramids) in (b) and (c). The resolution scales (red) are defined by factors $s_f$ of $1$ and $0.5$ with respect to the mapping resolution in (a), respectively. (g) and (h) illustrate maps resulting from fusing the data sequentially. The variable diffusion effects show that our method can accomodate noisy multiresolution measurements and capture spatial correlations.
   }\label{F:mapping_approach}
\end{figure}}

\section{Planning approach} \label{S:planning_approach}
This section details our planning scheme for terrain monitoring.
As depicted in \Cref{F:ipp_system},
we generate fixed-horizon plans
to maximize an objective.
To do this efficiently,
an evolutionary technique is applied to optimize trajectories
initialized by a 3-D grid search
in the \ac{UAV} workspace.
We first describe our approach to parameterizing trajectories,
then detail the algorithm itself.

\subsection{Trajectories}
A polynomial trajectory $\psi$ is represented
by a sequence of $N$ control waypoints to visit 
$\mathcal{C} = \{\mathbf{c}_1, \ldots, \mathbf{c}_N\}$ connected using $N-1$ $k$-order spline segments.
Given a reference velocity and acceleration,
we optimize the trajectory for smooth minimum-snap dynamics~\citep{Richter2013}.
The first waypoint $\mathbf{c}_1$ is clamped as the initial \ac{UAV} position.
As discussed in \Cref{S:problem_statement},
the function \textsc{measure(\textperiodcentered)} in \Cref{E:ipp_problem}
is defined by computing the spacing of measurement sites
along $\psi$ given a constant sensor frequency
and the traveling speed of the \ac{UAV} along the trajectory.

\subsection{Algorithm}
A fixed-horizon approach is used to plan adaptively
as data \editrev{is} collected.
During a mission, we alternate between replanning and execution
until the elapsed time $t$ exceeds a budget $B$.
Our replanning approach consists of two stages
and is shown in \Cref{A:replan_path}.
First, an initial trajectory, defined by $N$ fixed control waypoints $\mathcal{C}$,
is derived through a coarse grid search (Lines~3-7) in the 3-D workspace.
This step proceeds sequentially, selecting points in a greedy manner
\edit{based on a chosen utility function $\mathrm{I}($\textperiodcentered$)$},
so that a rough solution can quickly be obtained.
Then, the trajectory is refined \edit{using \Cref{E:ipp_problem}} to maximize the information objective.
In this step, we employ a generic evolutionary routine (Line~8)
\edit{to optimize the set of control waypoints $\mathcal{C}$}.

In \Cref{A:replan_path},
$\mathcal{Z}$ symbolizes a general model of the environment $\mathcal{E}$
capturing either a discrete or continuous target variable of interest.
\edit{The following sections
outline the key steps of the algorithm
before discussing possible ways of defining the objective for informative planning,
including in scenarios with adaptivity requirements \editrev{that require learning and focusing on objects or regions of interest online as they are discovered}.}
\begin{algorithm}[!h]
\Crefname{equation}{Eq.}{Equations}
\renewcommand{\algorithmicrequire}{\textbf{Input:}}
\renewcommand{\algorithmicensure}{\textbf{Output:}}
\algrenewcommand\algorithmiccomment[2][\scriptsize]{{#1\hfill\(//\)
\textcolor[rgb]{0.4, 0.4, 0.4}{#2} }}
\begin{algorithmic}[1]
  \Require Current environment model $\mathcal{Z}$,
  \edit{initial position $\mathbf{c}_1$,}
  number of control waypoints $N$,
  lattice points $\mathcal{L}$
  \Ensure Waypoints defining next polynomial trajectory $\mathcal{C}$
  \State $\mathcal{Z}' \gets \mathcal{Z}$ \Comment{Create a local copy of the map.}
  \edit{\State $\mathcal{C} \gets \mathbf{c}_1$; $\mathbf{c}_{\textrm{prev}} \gets \mathbf{c}_1$}
    \Comment{Initialize control points.}
  \While {$|\mathcal{C}| \leq N$}
  \State \edit{$\mathbf{c}^* \gets \underset{\mathbf{c} \in \mathcal{L}}{\argmax} 
    \frac{\mathrm{I}(\mathbf{c})}{\textsc{cost}(\mathbf{c},\,\mathbf{c}_{\textrm{prev}})}$
  \Comment{Find next-best point.}}
  \State $\mathcal{Z}' \gets$ \Call{predict\_measurement}{$\mathcal{Z}'$, $\mathbf{c}^*$}
    \Comment{From this point.}
  \State $\mathcal{C} \gets \mathcal{C} \cup \mathbf{c}^*$
    \Comment{Add to trajectory solution.}
  \State $\mathbf{c}_{\textrm{prev}} \gets \mathbf{c}^*$
  \EndWhile
  \State $\mathcal{C} \gets$ \Call{cmaes}{$\mathcal{C}$, $\mathcal{Z}$}
\Comment{Optimize control points using \Cref{E:ipp_problem}.}
\end{algorithmic}
\caption{\textsc{replan\_path} procedure}\label{A:replan_path}
\end{algorithm}

\subsection{3-D grid search} \label{SS:grid_search}
\begin{figure}[h]
  \centering
  \begin{subfigure}[]{0.49\columnwidth}
  \includegraphics[width=\columnwidth]{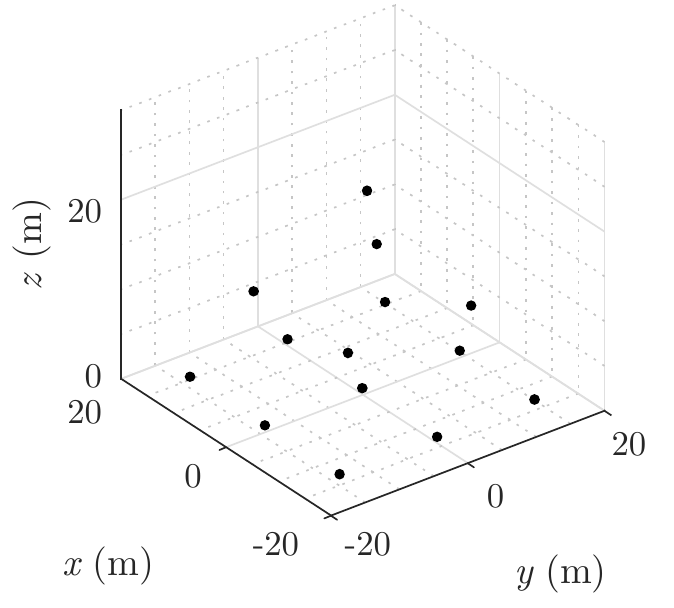}
  \caption{} \label{SF:lattice14}
  \end{subfigure}
  \begin{subfigure}[]{0.49\columnwidth}
  \includegraphics[width=\columnwidth]{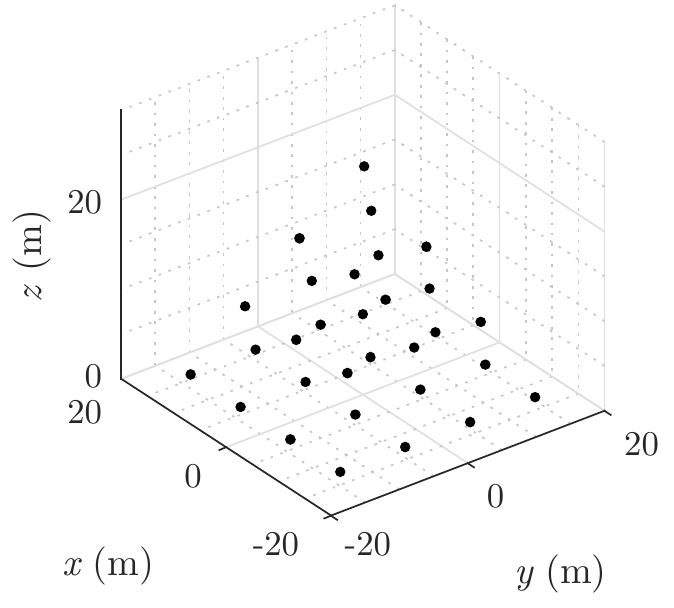}
  \caption{} \label{SF:lattice30}
  \end{subfigure}
   \caption{Visualizations of (a) 14-point and (b) 30-point
   3-D lattice grids $\mathcal{L}$ for obtaining an initial trajectory solution
   in a $40$\,m$\times$\,$40$\,m$\times$\,$30$\,m space.
   The point density can be chosen to trade off between solution accuracy and computational efficiency in the grid search.
   Note that the points are sparser at the top
   due to increasing \ac{FoV}.}\label{F:lattice}
\end{figure}

The first step of the replanning procedure (Lines~3-7 of \Cref{A:replan_path})
supplies an initial solution for the optimization step in \Cref{SS:optimization}.
To achieve this, the planner performs a 3-D grid search
based on a coarse multiresolution lattice $\mathcal{L}$
in the robot workspace (\Cref{F:lattice}).
A low-accuracy solution neglecting sensor dynamics is obtained efficiently
by using the points in $\mathcal{L}$ to represent candidate measurement sites
and assuming constant velocity travel.
Unlike in frontier-based exploration commonly used in cluttered environments~\citep{Charrow2015},
selecting goal measurement sites based on map boundaries is not applicable
in our setup.
Instead, we conduct a sequential greedy search for $N$ waypoints (Line~3),
where the next-best point $\mathbf{c}^*$ (Line~4)
is found by evaluating \edit{the \editrev{incremental} information gain rate given}
the chosen utility definition $\mathrm{I}($\textperiodcentered$)$ over $\mathcal{L}$.
\edit{This term represents the information objective and is discussed in detail in \Cref{SS:utility}. 
Since the number of waypoints $N$ is specified as a constant,
note that our approach in this step closely resembles
\editrev{the sequential greedy algorithm~\citep{Krause2008}
and generalized cost-benefit greedy algorithm~\citep{Zhang2016}}
for submodular function optimization with cardinality constraints,
with the requirement that travel time of the output trajectory defined by the points
must lie within the budget $B$.}
\editrev{However,
the optimization subproblem in \Cref{A:replan_path} fundamentally differs from cardinality constrained submodular maximization
in that our objective function is \textit{not} submodular.}

\edit{Importantly,
the prediction step assumes that no prior knowledge about the value of future measurements is available.
The search is thus conditioned on the \emph{most likely estimate} of the current field map,
i.e., considering that the maximum probability distribution of the map state will be re-observed.
For discrete mapping,
this involves partitioning the occupancy grid cells with
$p(\mathbf{x}_i) \geq 0.5$ as being occupied and those with $p(\mathbf{x}_i) < 0.5$ as being free
in order to identify which of the two classification curves in \Cref{F:sensor_binary} 
to apply for the map update procedure (\Cref{E:occupancy_grid_update}).
For continuous variable mapping,
the most likely estimate of the map simply corresponds to the current mean $\boldsymbol{\upmu}$ of the field distribution
and the update is performed using the Bayesian data fusion technique (\Cref{E:kf_mean,E:kf_cov}).}

For each $\mathbf{c}^*$, a fused measurement is simulated in $\mathcal{Z}$
\edit{using the relevant update strategy} (Line~5).
This point is then added to the initial trajectory solution (Line~6).

As depicted in \Cref{F:lattice},
the length scales of $\mathcal{L}$ can be defined
based on the computational resources available
and the level of accuracy desired;
the denser grid in \Cref{SF:lattice30} procures better initial solutions
at the expense of longer evaluation times.

\subsection{Optimization} \label{SS:optimization}
The second step (Line~8 of \Cref{A:replan_path}) optimizes the coarse grid search solution for \edit{the control waypoints} $\mathcal{C}$ \edit{using \Cref{E:ipp_problem}.
This is done by computing} $\mathrm{I}($\textperiodcentered$)$
for a sequence of measurements taken along the trajectory,
as defined in \Cref{SS:utility}.
Thereby,
\edit{an informed initialization procedure is used to speed up the convergence of the optimizer}.
Note that this step is agnostic to the optimization method considered;
in our specific approach,
we apply the Covariance Matrix Adaptation Evolution Strategy (CMA-ES),
discussed below.
In \Cref{SSS:optimization_comparison},
we evaluate our choice by comparing the use of different routines.

The CMA-ES is a generic global optimization routine
based on the concepts of evolutionary algorithms
which has been successfully applied to high-dimensional,
nonlinear, non-convex problems
in the continuous domain.
As an evolutionary strategy,
the CMA-ES operates by iteratively sampling candidate solutions
according to a multivariate Gaussian distribution in the search space.
Further details,
including a convergence analysis,
are provided in the in-depth review by~\citet{Hansen2006}.

Our choice of optimization method is motivated
by the nonlinearity of the objective space in \Cref{E:ipp_problem}
as well as by previous results~\citep{Popovic2017ICRA,Popovic2017IROS,Hitz2015}.
We initialize the mean solution with the grid search result,
constraining $\mathbf{c}_1$ to coincide with the current robot pose,
as described above.
In addition,
a coordinate-wise boundary handling algorithm~\citep{Hansen2009}
is applied to constrain the measurement points to lie
within a feasible cubical volume
of the workspace above the terrain.
For optimization,
we use the strategy internal parameters proposed by~\citet{Hansen2006},
and tune only the population size (number of offspring in the search),
co-ordinate wise initial standard deviation (step size, representing the distribution spread),
and maximum number of iterations
based on application-specific requirements.

\subsection{Utility definition} \label{SS:utility}
The utility, or information gain, function $\mathrm{I}($\textperiodcentered$)$ is critical
as it encapsulates the specific interests
for data-driven planning (\Cref{E:ipp_problem}).
This section discusses possible ways of quantifying
the value of new sensor measurements
with respect to the proposed map representations.
\editrev{We first introduce utility functions for a pure exploration scenario,
in which information gain depends only on the map uncertainty.
\Cref{SSS:adaptivity_requirements} then discusses objectives
for missions with adaptivity requirements,
where the aim is to discover and focus on targeted regions of interest in the environment}.

We examine definitions of $\mathrm{I}($\textperiodcentered$)$ for evaluating the exploratory value
of a measurement from a pose $\mathbf{p}$ (Line~4 of \Cref{A:replan_path}).
Note that, above,
$\mathbf{c}$ denotes a control waypoint parameterizing a polynomial trajectory,
whereas here $\mathbf{p}$ is a generic pose from where the measurement is registered.
In particular,
\edit{in a pure non-targeted exploration scenario},
we consider maximizing the reduction of Shannon's entropy $\mathrm{H}$
in the map $\mathcal{X}$:
\begin{equation}
  \mathrm{I}(\mathbf{p}) = \mathrm{H}(\mathcal{X}^-) - \mathrm{H}(\mathcal{X}^+) \, \textrm{,}
 \label{E:info_objective}
\end{equation}
where the superscripts denote the prior and posterior maps
given a measurement taken from $\mathbf{p}$.

In the discrete variable scenario,
the value of $\mathrm{H}$ for the occupancy map $\mathcal{X}$
is obtained by simply summing over the entropy values of all cells $\mathbf{x}_i \in \mathcal{X}$,
assuming their independence:
\begin{equation}
\begin{aligned}
  \mathrm{H}(\mathcal{X})& = \\
  & -\sum_{\mathbf{x}_i \in \mathcal{X}} p(\mathbf{x}_i)\log{p(\mathbf{x}_i)} + (1 - p(\mathbf{x}_i))\log{(1-p(\mathbf{x}_i))} \, \textrm{,} \label{E:discrete_info_objective}
\end{aligned}
\end{equation}
where $p(\mathbf{x}_i)$ indicates the probability of occupancy at $\mathbf{x}_i$.

In the continuous variable scenario,
however,
calculating $\mathrm{H}$
involves the determinant of the covariance matrix $\mathbf{P}$ of the \ac{GP} model~\citep{Rasmussen2006}.
We avoid this computationally expensive step
by instead maximizing the decrease in the matrix trace,
\edit{which is equivalent to minimizing the criterion for A-optimal design},
and only measures the total variance of the map cells~\citep{Sim2005}:
\begin{equation}
 \mathrm{I}(\mathbf{p}) ={} \Tr(\mathbf{P}^-) - \Tr(\mathbf{P}^+) \, \textrm{,} \label{E:continuous_info_objective}
\end{equation}
where $\Tr($\textperiodcentered$)$ denotes the trace of a matrix.

Note that \Cref{E:info_objective}
defines $\mathrm{I}($\textperiodcentered$)$ for a single measurement site $\mathbf{p}$.
To determine the utility of a complete trajectory $\psi$,
the same principles can be applied
by fusing a sequence of measurements \edit{iteratively} and computing the overall information gain.

\edit{\subsubsection{\editrev{Adaptivity for regions of interest}} \label{SSS:adaptivity_requirements}}
We also study an adaptive planning setup
\edit{where the aim is to map targeted objects or areas of interest,
such that} the objective depends on the values of the measurements taken
in addition to their location.
This property is very valuable for practical monitoring applications,
such as finding function extrema~\citep{Marchant2014},
classifying level sets~\citep{Hitz2014},
or focusing on specific value ranges.
To this end,
\Cref{E:info_objective} is modified
so that the elements mapping to the value of each cell
$\mathbf{x}_i \in \mathcal{X}$
are excluded from the objective computation,
provided they do not satisfy the requirement which defines interest-based planning.

As an example,
\editrev{this work considers a mission where the aim is to focus specifically on regions of interest which have higher values of the latent target parameter,
e.g., areas of high vegetation cover in a field.
A threshold is applied
to separate the (a) ``interesting" (above) and (b) ``uninteresting" (below) parameter value range
into complementary subsets (a) $\mathcal{X}_I$ and (b) $\mathcal{X}_-$ of all points $\mathcal{X}$ in the map.
The partitioning strategy is described below.
The main idea is to selectively include only the points $\mathcal{X}_I$ in calculating the information value of potential measurements in \Cref{E:info_objective},
i.e., the utility associated with lower values in the points $\mathcal{X}_{-} = \mathcal{X} \setminus \mathcal{X}_{I}$ is ignored.}

In the discrete case, this simply amounts to computing \Cref{E:info_objective}
\editrev{only} for the upper set of interesting cells
$\mathcal{X}_I = \{\mathbf{x}_i \in \mathcal{X}\,|\,p_{th} < p(\mathbf{x}_i)\}$,
where $p_{th}$ is a threshold on probabilistic occupancy.
However,
\editrev{in the continuous case},
to account for model uncertainty when planning with the \ac{GP} model (\Cref{E:continuous_info_objective}),
we adopt the principles of bounded uncertainty-aware classification from \citep{Gotovos2013,Srinivas2012}.
The subset of interesting locations $\mathcal{X}_I$
is defined based on the mean and variance of each cell $(\mu_i, \sigma_i)$ as:
\begin{equation}
 \mathcal{X}_I = \{ \mathbf{x}_i\, |\,
  \mathbf{x}_i \in \mathcal{X} \wedge
   \mu_i + \beta\sigma_i \geq \mu_{th} \} \, \textrm{,} \label{E:adaptive_planning}
\end{equation}
where $\mu_{th}$ is a threshold on the underlying scalar field
and $\beta$ is a design parameter tuned to scale the confidence interval for classification,
i.e., the certainty below $\mu_{th}$ a cell must possess before being considered interesting.

\vspace{4mm}

\section{Experimental results} \label{S:results}

This section discusses our experimental findings
in both continuous (\Cref{SS:simulation_evaluation}) and discrete (\Cref{SS:dataset_evaluation})
mapping scenarios.
\editrev{The overall aim is to assess the performance of our framework and demonstrate its flexibility
to cater for both types of scenario in missions with and without requirements to adapt to targeted regions of interest.}
First, in \Cref{SS:simulation_evaluation},
we evaluate the performance of the proposed approach in simulation
and examine the influence of its key parameters.
For these experiments,
we consider the more complex case of mapping a continuous target variable
in a bounded environment
using a \ac{UAV} equipped with an image-based classifier.
\Cref{SSS:benchmark_comparison} and \Cref{SSS:optimization_comparison}
compare our approach to state-of-the-art methods
and study the effects of using different optimization routines
in our algorithm.
The adaptive replanning scheme
is evaluated in \Cref{SSS:adaptive_replanning}.
Then, in \Cref{SS:dataset_evaluation},
we demonstrate the application of our framework
for a realistic active classification problem
using the RIT-18 dataset~\citep{Kemker2018},
\edit{before validating real-time capabilities
in a real world agricultural monitoring scenario.}
\begin{figure*}[!h]
\centering
\begin{subfigure}{.24\textwidth}
\includegraphics[width=\textwidth]{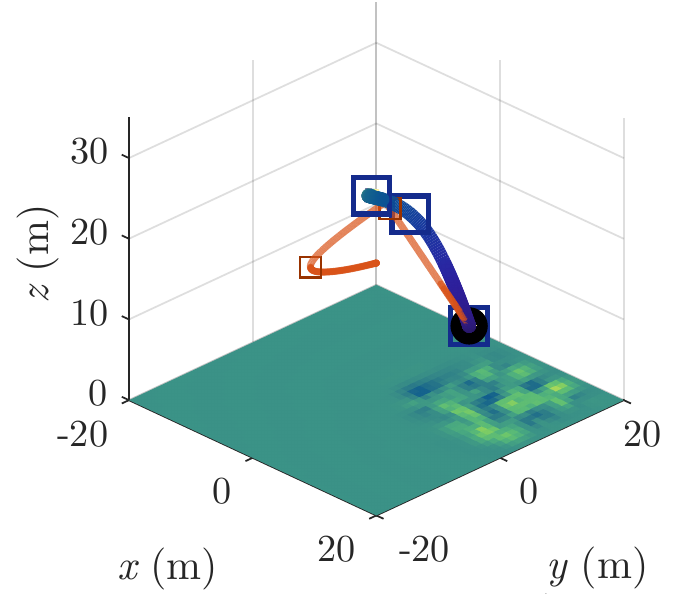}%
\end{subfigure}\hfill%
\begin{subfigure}{.24\textwidth}
\includegraphics[width=\textwidth]{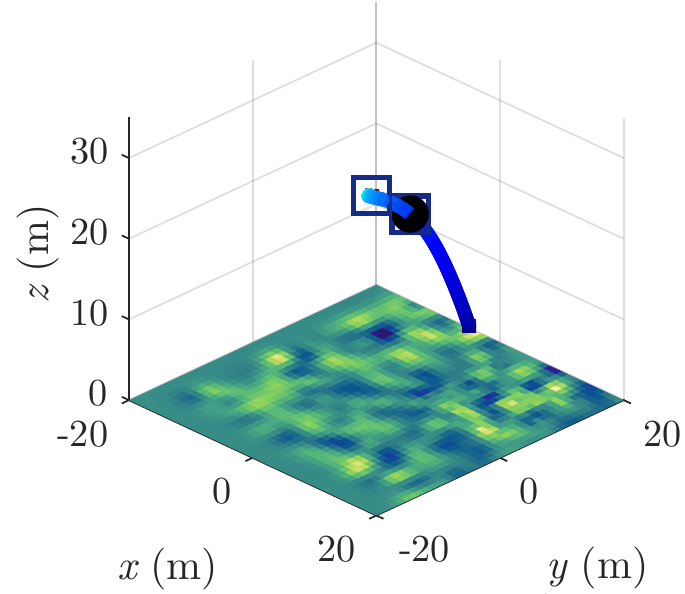}%
\end{subfigure}\hfill%
\begin{subfigure}{.24\textwidth}
\includegraphics[width=\textwidth]{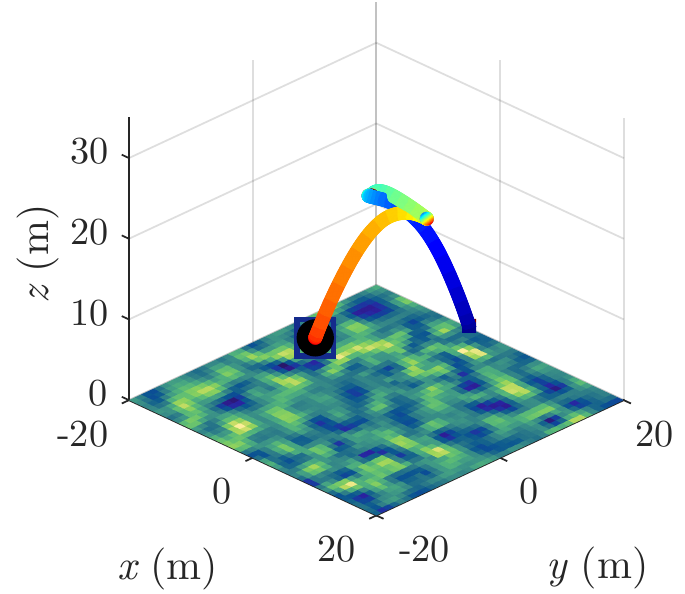}%
\end{subfigure}\hfill%
\begin{subfigure}{.24\textwidth}
\centering
\includegraphics[width=0.95\textwidth]{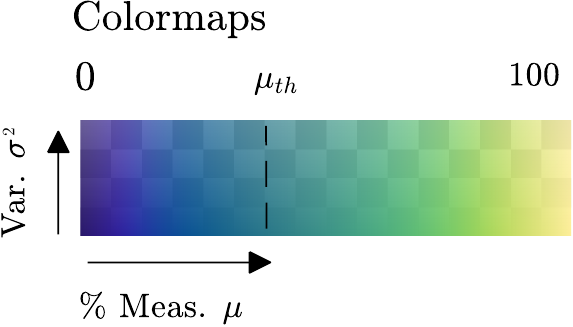} \\
\vspace{1.5mm}
\includegraphics[width=0.9\textwidth,right]{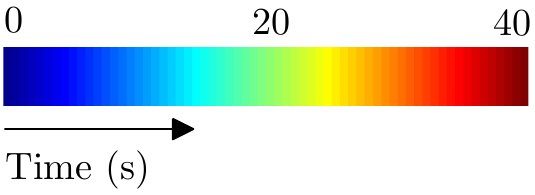}%
\end{subfigure}%
\vspace*{1mm}
\begin{subfigure}[]{.165\textwidth}
\includegraphics[width=\textwidth]{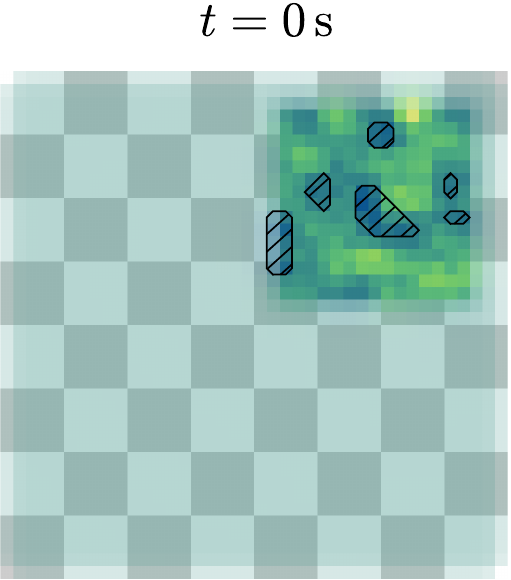}%
\end{subfigure}%
\hspace*{14mm}
\begin{subfigure}{.165\textwidth}
\includegraphics[width=\textwidth]{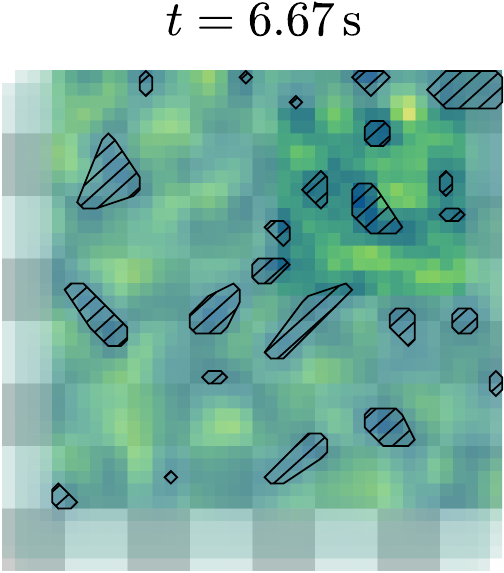}%
\end{subfigure}%
\hspace*{16mm}
\begin{subfigure}{.165\textwidth}
\includegraphics[width=\textwidth]{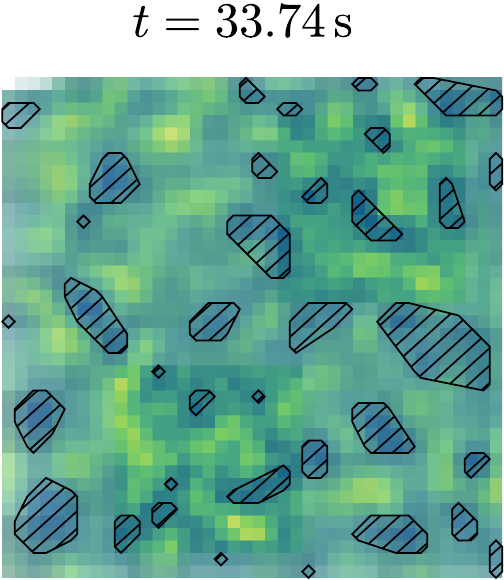}%
\end{subfigure}%
\hspace*{5mm}
\begin{minipage}{.195\textwidth}
\vspace*{5mm}
\hspace*{3.5mm}
\includegraphics[width=\textwidth]{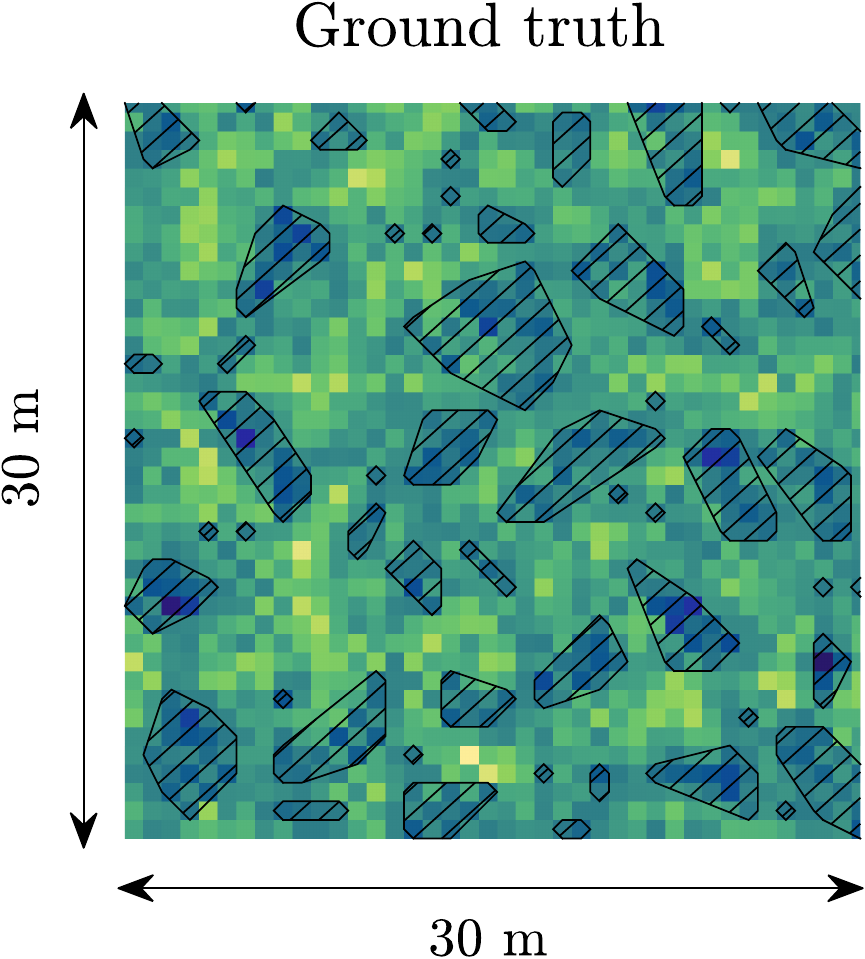}%
\end{minipage}%
\caption{Example simulation results of our \ac{IPP} framework. The colormaps are shown on the top-right.
Bluer and yellower shades represent lower and higher values of the target parameter, respectively.
In the bottom maps, opacity indicates the model uncertainty (variance, $\sigma_i^2$),
with the checkerboard added for visual clarity and the hatched sections denoting uninteresting areas with $< \mu = 40\%$.
The ground truth is shown on the bottom-right.
The three columns on the left depict the trajectories (top row) and maps (bottom row) at
different snapshots of the mission at times $t = 0$\,s, $6.67$\,s, and $33.74$\,s.
In the top plots, the black dot indicates the current \ac{UAV} position while the squares show the measurement sites.
The top-left figure illustrates an example trajectory before (orange) and after (colored gradient) optimization using the CMA-ES.
Note that the map means are rendered in the top trajectory plots.}
\label{F:example}
\end{figure*}

To begin, \Cref{F:example} presents
an illustrative example of the progression of our framework
for mapping an \textrm{a priori} unknown environment.
For adaptive planning, we set a base threshold $\mu_{th} = 40\%$ \edit{in \Cref{E:adaptive_planning}}
to focus on the more interesting, higher-valued target parameter range.
This value allows us to also include unobserved cells in the objective,
which are initialized uniformly with a uninformed mean prior of $50\%$.
The top and bottom rows visualize the planned \ac{UAV} trajectories and maps, respectively,
as images are registered at different times during the mission.
The top-left plot depicts the first planned trajectory before (orange) and after (colored gradient) applying the CMA-ES.
As shown, the optimization step shifts initial measurement sites (squares) to high altitudes,
allowing low-resolution, high-uncertainty data to be quickly collected
before the map is refined (second and third columns).
A qualitative comparison with ground truth (bottom-right) confirms that our method performs well,
producing a a fairly complete map in a short time period
with most uninteresting regions (hatched areas) identified.

\subsection{Simulation results} \label{SS:simulation_evaluation}

\subsubsection{Comparison against benchmarks} \label{SSS:benchmark_comparison}
Our framework is evaluated by comparison to benchmarks
for continuous variable mapping in a simulated environment.
\edit{The simulations were run in MATLAB
on a single desktop with a $1.8$\,GHz Intel i7 processor and $16$\,GB of RAM}
and model a synthetic information gathering problem
in a $30$\,m$\times$\,$30$\,m area.
The target distributions are generated
as 2-D Gaussian random fields
with the mapped scalar parameter ranging from $0$\,\% to $100$\,\%
and cluster radii randomly set between $1$\,m and $3$\,m.
We use a uniform resolution of $0.75$\,m for both the training $\mathcal{X}$ and predictive $\mathcal{X}^*$ grids,
and perform uninformed initialization with a uniform mean prior of $50$\,\%.
For the \ac{GP}, an isotropic Mat\'ern 3/2 kernel is applied.
It is defined as~\citep{Rasmussen2006}:
\begin{equation}
  k_{Mat3}(\mathbf{x},\mathbf{x}') =
  \sigma^2_f\,(1+ \frac{\sqrt{3}d}{l})\,\exp{(-\frac{\sqrt{3}d}{l})} \, \textrm{,} \label{E:matern_kernel}
\end{equation}
where $d$ is the Euclidean distance between inputs $\mathbf{x}$ and $\mathbf{x}'$,
and $l$ and $\sigma_f^2$ are hyperparameters representing the length scale and signal variance, respectively.
We train the hyperparameters $\{\sigma_n^2,\,\sigma_f^2,\,l\} = \{1.42, \, 1.82, \, 3.67\}$
by maximizing the model log marginal likelihood,
using four independent maps with variances modified
to cover the entire target parameter range during inference.

For fusing new data,
measurement noise is simulated based on the camera model in \Cref{F:sensor_continuous},
with a $10$\,m altitude beyond which images scale by a factor of $s_f = 0.5$.
This places a realistic limit on the quality of data that can be obtained from higher altitudes.
We consider a square camera footprint
with $60\degree$ \ac{FoV} and a $0.15$\,Hz measurement frequency.
For the purposes of these experiments,
we assume no actuation or localization noise
and that the on-board camera always faces downwards.

Our approach is compared against three different strategies:
(a) the sampling-based rapidly exploring information gathering tree (RIG-tree)
introduced by~\citet{Hollinger2014}, a state-of-the-art \ac{IPP} method;
(b) a traditional \edit{fixed-altitude} ``lawnmower'' coverage path;
\edit{(c) an upward spiral 3-D coverage path}; and
(d) random waypoint selection.
\edit{The random approach is considered
as a na\"{i}ve benchmark that does not incorporate \editrev{either} the structure \editrev{or} the state of the field map for planning,
but can be easily implemented in practice.}
A $200$\,s budget $B$ is allocated for all strategies.
Considering that, to the best of our knowledge,
there is no \ac{IPP} method that procures optimal results
when operating in the continuous trajectory space,
we assess performance by comparing different information metrics
during a mission.
We quantify uncertainty with the trace of the map covariance matrix $\Tr(\mathbf{P})$
and study the \ac{RMSE} and \ac{MLL} at points in $\mathcal{X}$ with respect to ground truth as accuracy statistics.
As described by~\citet{Marchant2014}, the \ac{MLL} is a probabilistic confidence measure
incorporating the variance of the predictive distribution.
Intuitively, all metrics are expected to reduce as data are acquired over time,
with steeper declines signifying better performance.

\edit{For our planner and methods (a), (b), and (d)},
the \ac{UAV} starting position is specified
\edit{in the corner of the environment}
as ($7.5$\,m,\,$7.5$\,m) within the field and $8.66$\,m altitude
to assert the same initial conditions as for the complete \edit{``lawnmower''} coverage pattern.
For trajectory optimization,
the maximum reference velocity and acceleration are $5$\,m$/$s and $2$\,m$/$s$^2$
using polynomials of order $k = 12$,
and the number of measurements along a path is limited to $10$ for computational feasibility.
In our planner, we define polynomials with $N = 5$ waypoints and use the lattice in \Cref{SF:lattice30} for the 3-D grid search.
In RIG-tree, we associate control waypoints with vertices,
and form polynomials by tracing the parents of leaf vertices to the root.
For both planners,
we consider the utility $\mathrm{I}($\textperiodcentered$)$ in \Cref{E:continuous_info_objective}
and set a base threshold of $\mu_{th} = 40\%$ \edit{in \Cref{E:adaptive_planning}}
above which map regions are considered interesting
\edit{to define an adaptive planning requirement}.

As outlined in our previous papers~\citep{Popovic2017ICRA,Popovic2017IROS},
we use a fixed-horizon version of RIG-tree
\edit{to allow for incremental replanning and adaptivity
and obtain a fair comparison to our planner.
We set the \ac{UAV} starting position as the same corner of the environment as described above,
with no prior map information used to create the initial plan.
The algorithm alternates between tree construction (replanning) and plan execution,
updating the map with each new set of measurements.
For replanning,
the tree branch expansion step size is set to $10$\,m
with $500$ sampling iterations.
The latter value was set to obtain the same $\sim 20$\,s replanning time
as required by the CMA-ES to optimize a single trajectory,
given $45$ iterations and the optimization paremeters suggested by~\citet{Hansen2006}.
This enables us to compare the methods fairly
in the experimental setup
based on the time taken to find an informative trajectory.}

In the \edit{fixed-altitude} coverage planner, height ($8.66$\,m) and velocity ($0.78$\,m$/$s) are defined
for complete \edit{uniform} coverage given the specified budget and measurement frequency.
To design a fair benchmark, we studied possible ``lawnmower'' patterns
with heights determined by the camera \ac{FoV}.
For each pattern, we modified velocity to match the budget,
then selected the best-performing one.
\edit{In the 3-D coverage planner,
the path is a conical spiral trajectory,
reducing in radius with height and
spanning the minimum ($1$\,m) and maximum ($26$\,m) \ac{UAV} flight altitudes.}
Finally, in the random planner,
we randomly sample a destination in the bounded volume above the terrain
and generate a trajectory by connecting it to the current \ac{UAV} position.

\begin{figure}[!h]
\centering
 \includegraphics[width=\columnwidth]{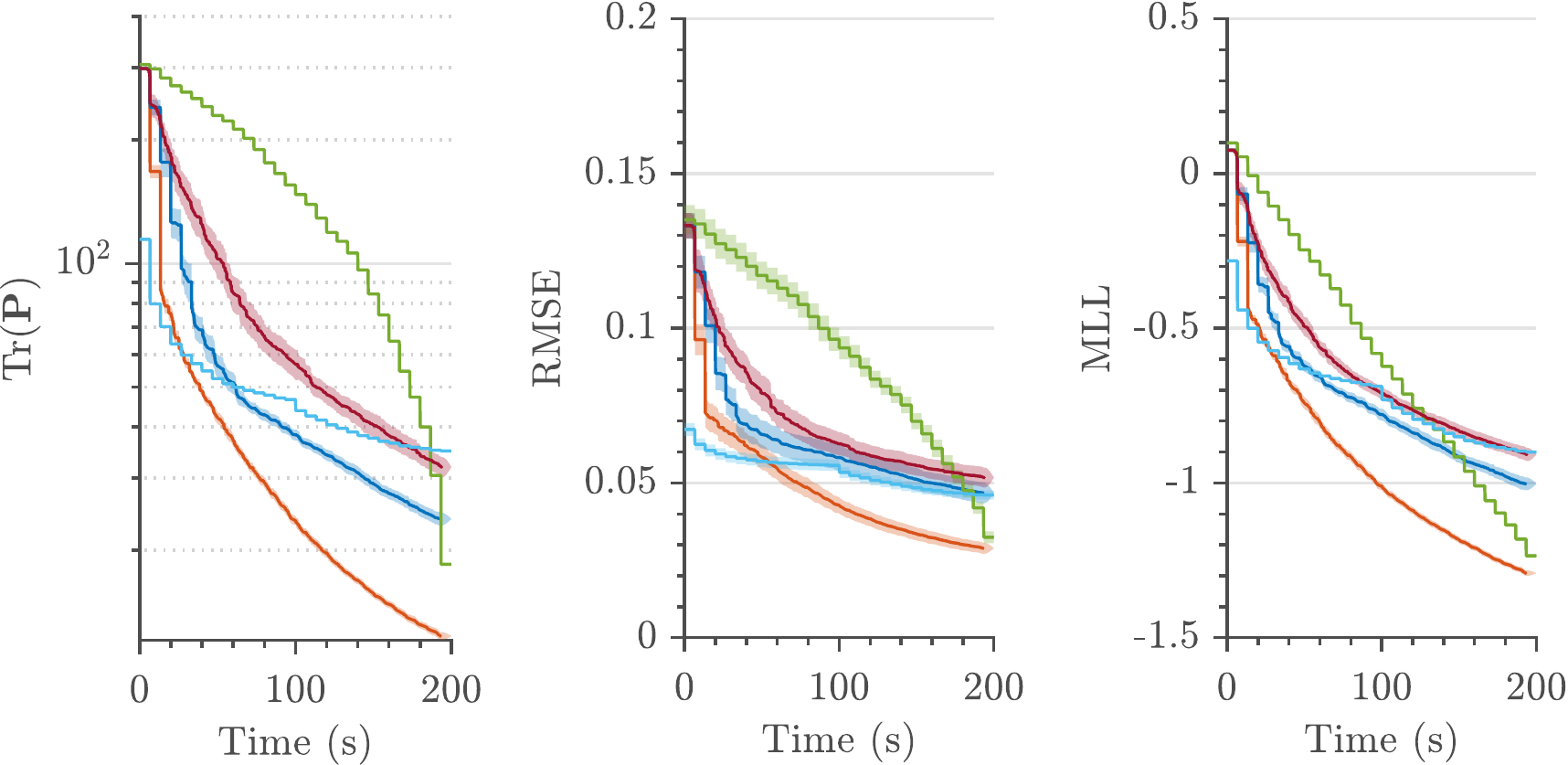} \\
 \vspace{2.5mm}
 \includegraphics[width=0.95\columnwidth]{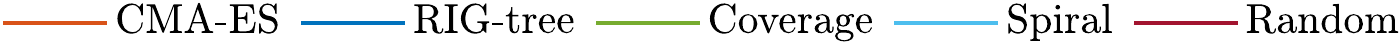}
 \caption{Comparison of our \ac{IPP} framework using the CMA-ES
 against benchmarks
 for a fixed mission time budget of $200$\,s.
 The solid lines represent means over 30 trials.
 The thin shaded regions depict 95\% confidence bounds.
 Using IPP, map uncertainty (left) and error (middle, right)
 reduce quickly as the \ac{UAV} obtains low-resolution images before descending.
 Note the logarithmic scale of the $\Tr(\mathbf{P})$ axis.} \label{F:methods}
\end{figure}
\Cref{F:methods} shows how the metrics evolve for each planner during the mission.
\edit{Note that the spiral curve (light blue) is offset at $t = 0$
since the location of the initial measurement is different.}
For our algorithm, we use the CMA-ES optimization method.
The \edit{``lawnmower''}  coverage curve (green) validates our previous results
that uncertainty (left) reduces uniformly and deterministically for a constant altitude and velocity.
\edit{This motivates approaches that permit the \ac{UAV} to fly at variable altitudes,
as they can} compromise between sensor uncertainty and \ac{FoV}.
\edit{By starting at the lowest altitude \editrev{before gradually ascending},
the spiral planner \editrev{(light blue)} improves the map most quickly \editrev{early on in} the missions
but levels off with the lower-quality measurements obtained as the \ac{UAV} \editrev{flies further up}.}
\editrev{In contrast,
with a fixed-size camera footprint,
the coverage performs better at later stages,
but is not capable of achieving the initial rapid map improvement.}
\edit{As expected,}
both our algorithm (light orange) and RIG-tree (\edit{dark} blue) perform better than
\edit{the spiral and random (dark red) benchmarks},
\editrev{given that} the latter do not \edit{exploit \ac{IPP} objectives} to guide selecting next waypoint destinations.

Our algorithm produces maps with lower uncertainty and error
than those of RIG-tree given the same budget.
This confirms that our two-stage planner is more effective than sampling-based methods
with the proposed mapping strategy.
We noted that fixed step size is a key drawback of RIG-tree,
because values allowing initial ascents tend to limit incremental navigation when later refining the map.

\begin{figure*}[!t]
\centering
\begin{minipage}{0.45\textwidth}
 \includegraphics[width=0.5\columnwidth]{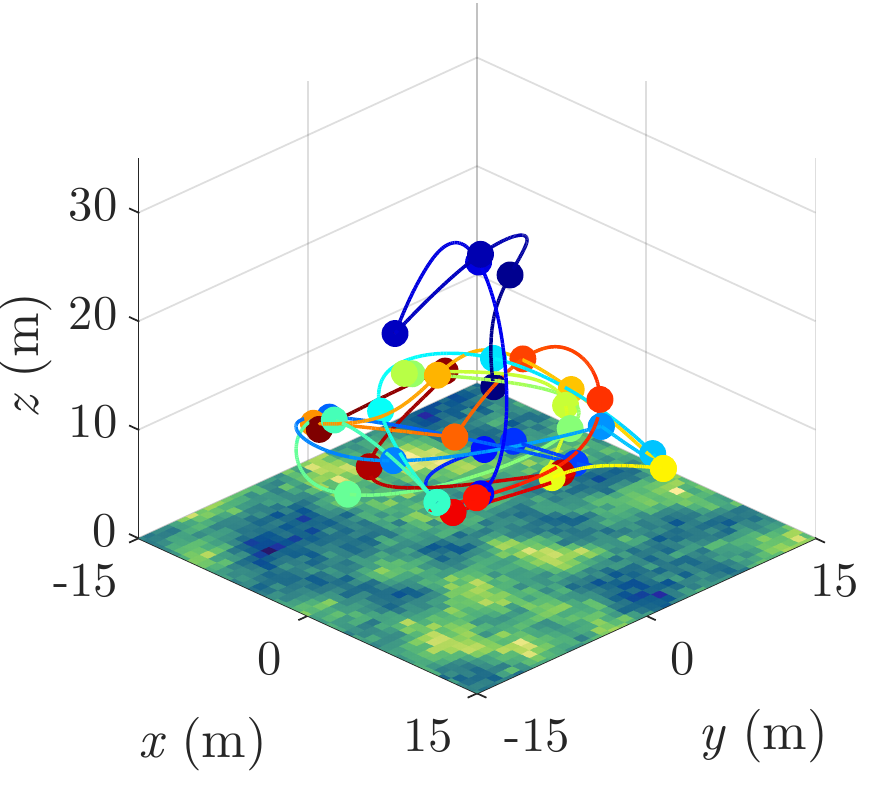}%
 \includegraphics[width=0.5\columnwidth]{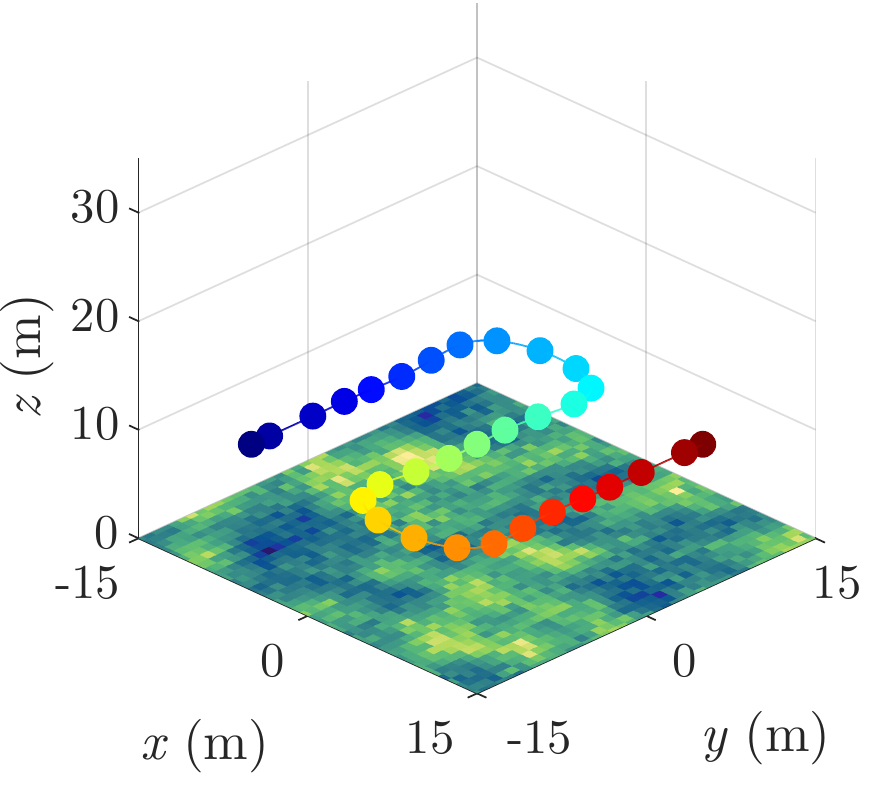}%
 \vspace{0.4cm}
 \includegraphics[width=\columnwidth]{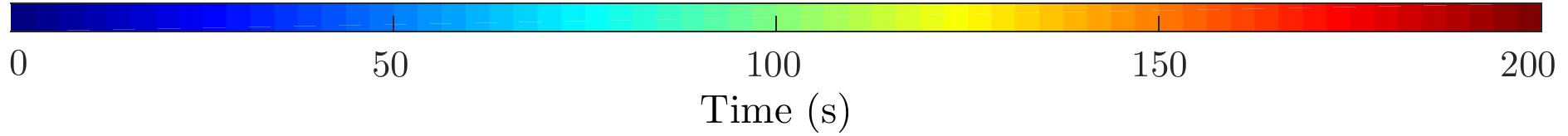}
 \end{minipage} %
 \hfill
 \begin{minipage}{0.54\textwidth}
 \includegraphics[width=0.5\columnwidth]{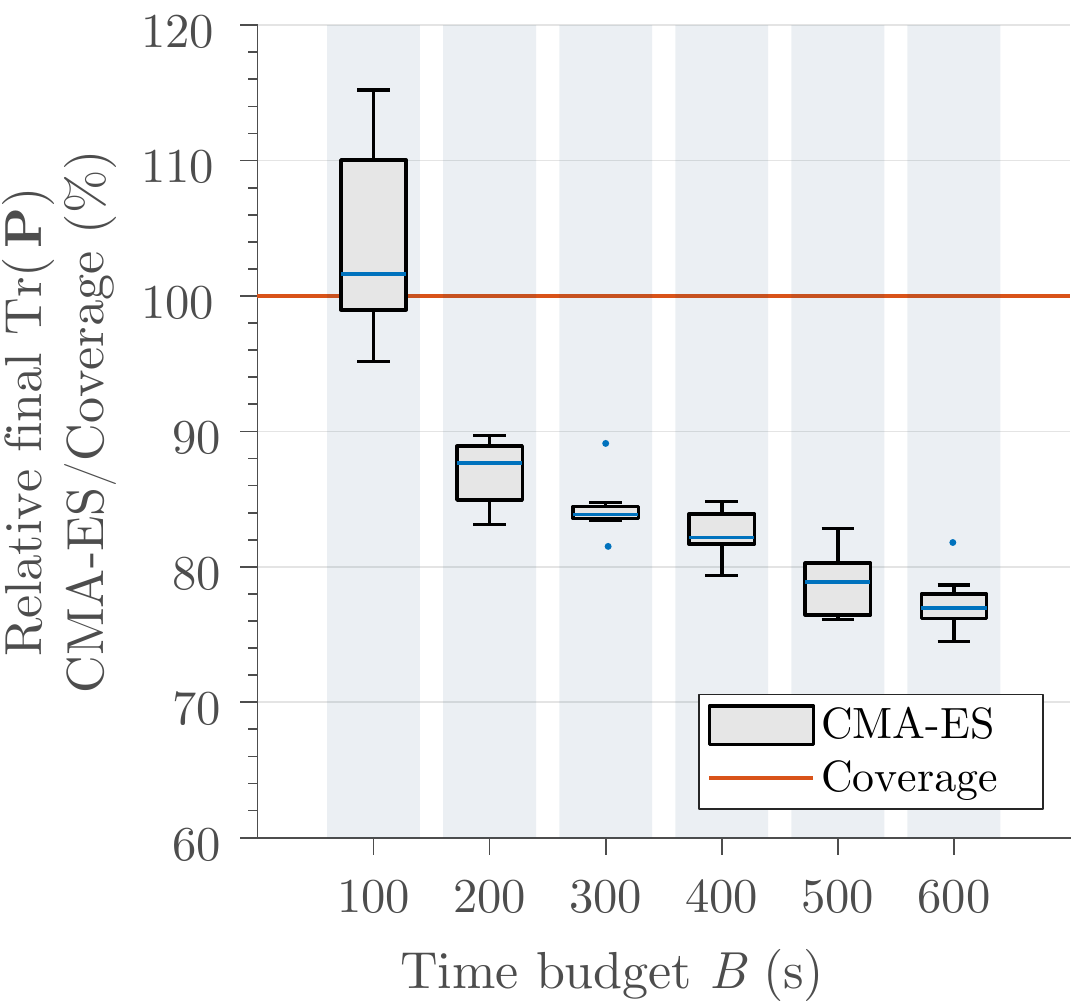}%
 \includegraphics[width=0.5\columnwidth]{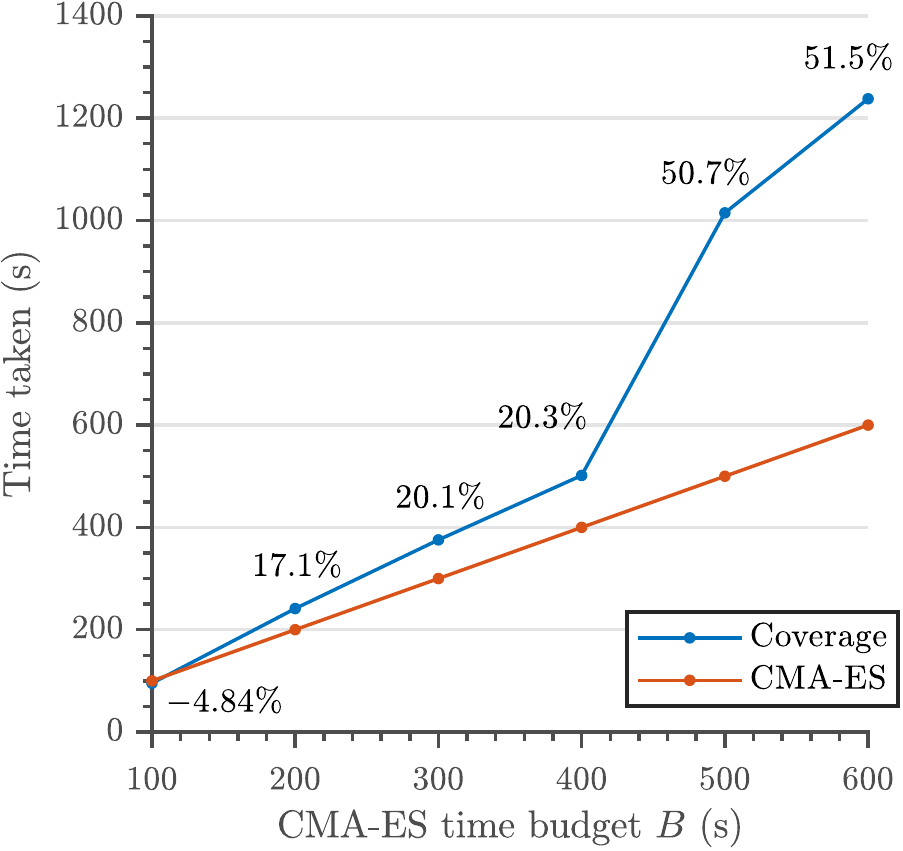}%
 \end{minipage}
 \caption{\emph{Left}:
 Example of an evaluation scenario
 comparing our IPP approach with the CMA-ES to ``lawnmower'' coverage
 (left and right plots, respectively) for mapping a continuous variable
 in $200$\,s missions.
 The colored lines represent the traveled trajectories,
 the spheres indicate measurement sites,
 and the ground truth maps are rendered.
 \emph{Middle}:
 Comparison of the final map uncertainties ($\Tr(\mathbf{P})$) for various path budgets.
 Ten CMA-ES trials were run for each budget.
 \emph{Right}:
 Comparison of times taken
 to achieve the same final map uncertainty,
 given a fixed path budget for the CMA-ES.
 \edit{The orange line corresponds to average values over $10$ CMA-ES trials,
 and relative time savings using the new method are shown as percentages.}
 By allowing for altitude variations,
 our approach trades off between \ac{FoV} and sensor noise
 to quickly obtain high-confidence maps
 with finer end quality
 in the same time period.} \label{F:coverage_comparison}
 \end{figure*}

Using the same simulation setup,
we also conducted a detailed comparison between our approach and ``lawnmower'' coverage
to examine the benefits of \ac{IPP} for missions of different durations.
First, we considered six budgets $B$
$(100$\,s,\,$200$\,s,\,\ldots\,, $600$\,s) \edit{on mission time.
For each budget, the proposed CMA-ES-based framework was tested over $10$ trials, giving a total of $60$ simulations,
and the coverage planner was run once with its deterministic path.
As detailed above, for a fair evaluation,
the fixed coverage altitude for each mission time
was chosen for best performance among different complete ``lawnmower'' patterns.}
As an example,
the left plots in \Cref{F:coverage_comparison} depict the trajectories executed
on a $200$\,s scenario used for the evaluation.
The middle graph shows a quantitative analysis
of the final achieved map uncertainties ($\Tr(\mathbf{P})$).
For comparison,
the results of our approach are normalized with the corresponding coverage planner value,
so that percentages below $100$\% (orange line) indicate a better performance of our method.
Second, on the right,
we compared the mission times needed by the two methods
to produce maps with the same final uncertainties.
\edit{In these experiments,
we examined the six fixed budgets for our method (orange)
and, for each,
investigated the time requirement for the coverage strategy (blue)
to obtain the same reconstruction quality}.

\Cref{F:coverage_comparison} illustrates
two key benefits of using our approach:
(a) we obtain maps with lower final uncertainty (middle)
by not fixing the flight altitude of the \ac{UAV},
and (b) we attain significant time savings (right)
by allowing for the early collection of low-quality data,
as evidenced in \Cref{F:methods}.
As a result, with increasing mission time,
the marginal discrepancy in the final maps increases.
The plot on the right shows that
our approach also requires substantially less time
($>50\%$ savings for $>500$\,s missions)
to achieve maps with the same uncertainty.
This is because the ``zig-zag'' pattern for these missions
must be set at a lower altitude ($6.5$\,m, compared to $8.66$\,m for lower budgets)
to obtain a reduced level of sensor noise,
which increases the total distance traveled by the \ac{UAV}.
Interestingly,
the coverage path produces a better result only at the end of a $100$\,s mission;
this is because, at this budget,
our planner lacks time to descend to refine the map.
In future studies,
we intend to extend our ideas from previous work~\citep{Popovic2017ICRA}
and include an awareness of the remaining path budget for planning.

\subsubsection{Comparison of optimization methods} \label{SSS:optimization_comparison}
Next,
we consider the effects of using different optimization routines in \Cref{SS:optimization}
to evaluate our CMA-ES-based approach.
We use the same simulation setup as d\textbf{}escribed above,
i.e., 30 trials for mapping a $30$\,m\,$\times$\,$30$\,m environment, 
and study the following methods:
\begin{enumerate}[label=(\alph*)]
 \item \emph{Lattice}: 3-D grid search only (i.e., without Line~7 in \Cref{A:replan_path});
 \item \emph{CMA-ES}: global evolutionary optimization routine~\citep{Hansen2006} (described in \Cref{SS:optimization});
 \item \emph{\ac{IP}}: approximate gradient-based optimization using the interior-point approach~\citep{Byrd2006};
 \item \emph{\ac{SA}}: global optimization based on the physical cooling process in metallurgy~\citep{Ingber1992};
 \item \emph{\ac{BO}}: global optimization using a \ac{GP} prior~\citep{Gelbart2014}.
\end{enumerate}
\edit{For optimizers (b)-(e)
we investigate initializations using:
(i) the lattice search output from (a)
and (ii) random point selection in the workspace,
in order to investigate their sensitivity to the starting conditions.}

The aim of these experiments is to examine how the methods compare
using standard \edit{MATLAB} implementations as baselines.
\edit{As described above,
we set the stopping criteria of the algorithms such that
each one is allowed the same $\sim 20$\,s for replanning.
Note that the benchmarks were applied 
using default or recommended values, without significant effort invested
into adjusting their parameters, in order to make them practically comparable with the CMA-ES,
which requires minimal tuning procedures.}

For the local \ac{IP} optimizer, we approximate Hessians by a dense quasi-Newton strategy
and apply the step-wise algorithm described by~\citet{Byrd2006}.
For \ac{SA},
we apply an exponential cooling schedule
and an initial temperature of $100$.
For \ac{BO}, we use
the time-weighted Expected Improvement acquisition function studied by~\citet{Gelbart2014}
with an exploration ratio of $0.5$.
Additionally,
\edit{for our approach},
we examine two variations of the CMA-ES
with initial step sizes of \edit{$(0.5$\,m,\,$0.5$\,m,\,$1$\,m$)$}, $(3$\,m,\,$3$\,m,\,$4$\,m$)$ and $(10$\,m,\,$10$\,m,\,$12$\,m$)$
in the $(x,y,z)$ co-ordinates,
where the $z$-axis defines altitude.
\edit{
These values were chosen based on the extent of the robot workspace
in order to compare different global search behaviors,
as the step size parameter effectively captures the distribution from which new solutions are sampled,
and thus how well the problem domain is covered by the optimization routine.
The aim is to also obtain a practical insight into 
the tuning requirements for the CMA-ES
to achieve best results.
}

\begin{figure}[!h]
\centering
 \includegraphics[width=0.98\columnwidth]{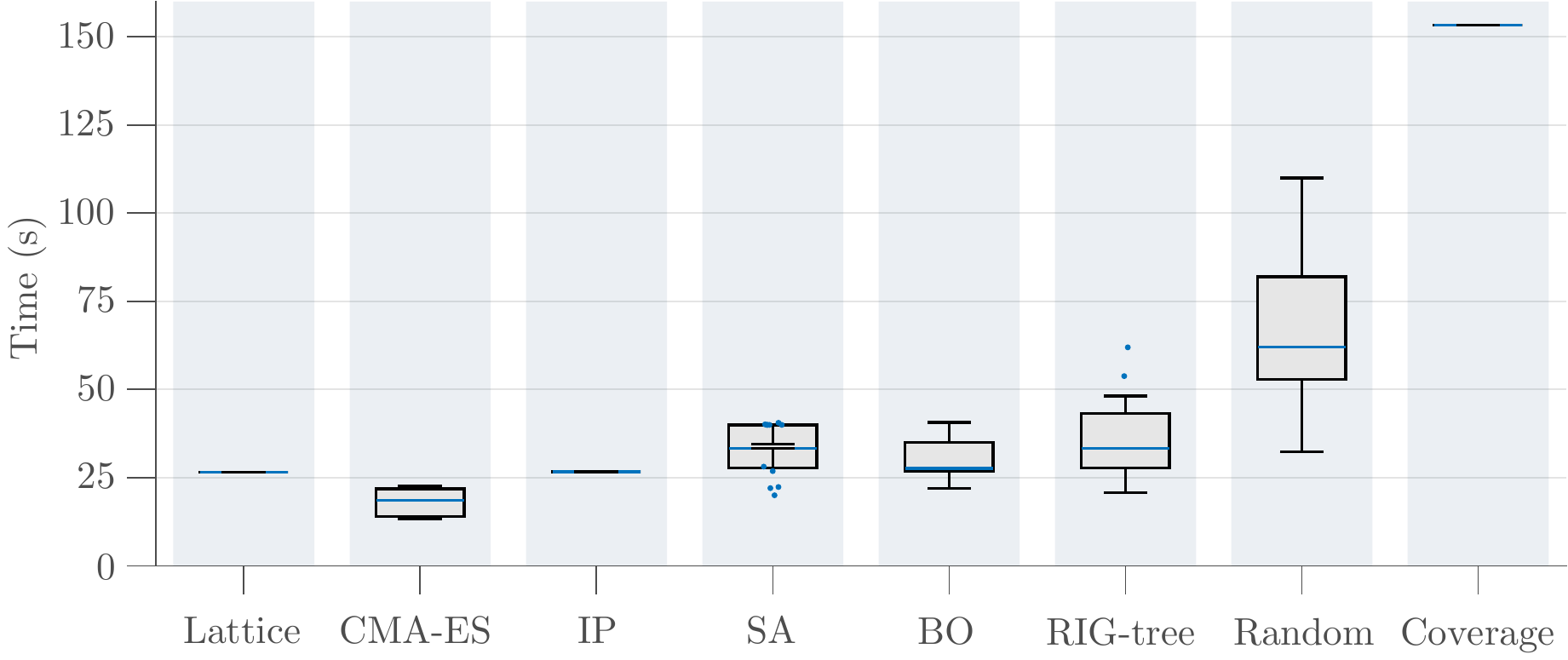}
 \caption{Mean times required by each optimization method to reduce $\Tr(\mathbf{P})$ (map uncertainty)
 to $75\%$ of its initial value, averaged over the $30$ trials.
 \edit{The CMA-ES, IP, SA, and BO routines use a lattice-based initialization strategy.}
 The CMA-ES result corresponds to a search with \editrev{intermediate} step sizes (`CMA-ES $(3,4)$').
 With a mean time of $18.1$\,s, this approach performs best.
 } \label{F:decay_times}
\end{figure}

\Cref{T:methods} displays the mean results for each method averaged over the $30$ trials,
with the benchmarks from \Cref{SSS:benchmark_comparison} included for reference.
\edit{The suffixes are used to denote the initialization strategy (lattice or random)
and different step sizes for the CMA-ES
\editrev{($(0.5,1)$, $(3,4)$, and $(10,12)$ correspond to step sizes of
$(0.5$\,m,\,$0.5$\,m,\,$1$\,m$)$, $(3$\,m,\,$3$\,m,\,$4$\,m$)$ and $(10$\,m,\,$10$\,m,\,$12$\,m$)$, respectively)}}.
Following~\citet{Marchant2014}, we also show weighted statistics to emphasize errors in high-valued regions.
As the same objective is used for all methods,
consistent trends are observed in both non-weighted and weighted metrics.
In \Cref{F:decay_times},
we show the mean times taken by each method \edit{using the lattice initialization}
to reduce $\Tr(\mathbf{P})$
to $75\%$ of its initial value
to represent the decay speed of the map uncertainty.
\begin{table}[!h]
\centering
\resizebox{\columnwidth}{!}{%
\begin{tabular}{lccccc} \toprule
    Method         & {$\Tr(\mathbf{P})$}      & {RMSE}    & {WRMSE}    & {MLL}     & {WMLL}  \\ \midrule \midrule
    Lattice        & 51.421          & 0.0560    & 0.0559     & -0.960    & -0.960  \\ \midrule
    \edit{CMA-ES (0.5,1) \edit{lat.}} & \edit{$\mathbf{45.670}$}   & \edit{$\mathbf{0.0520}$}   & \edit{$\mathbf{0.0517}$}     & \edit{$\mathbf{-1.007}$}    & \edit{$\mathbf{-1.009}$}  \\ \midrule
    CMA-ES (3,4) \edit{lat.}   & $\mathbf{45.187}$ & $\mathbf{0.0525}$    & $\mathbf{0.0522}$     & $\mathbf{-1.002}$    & $\mathbf{-1.006}$  \\ \midrule
    CMA-ES (10,12) \edit{lat.} & 51.246          & 0.0603    & 0.0599     & -0.894    & -0.897  \\ \midrule
    IP \edit{lat.}            & 47.973          & 0.0541    & 0.0538     & -0.961    & -0.962  \\ \midrule
    SA \edit{lat.}            & 51.968          & 0.0571    & 0.0569     & -0.903    & -0.904  \\ \midrule
    BO \edit{lat.}            & 68.806          & 0.0685    & 0.0682     & -0.747    & -0.749  \\ \midrule
    Random         & 92.618          & 0.0771    & 0.0764     & -0.668    & -0.669  \\ \midrule
    \edit{CMA-ES (3,4) \small{rand.}}  & \edit{53.448}         & \edit{0.0613}    & \edit{0.0608}     & \edit{-0.8614}    & \edit{-0.8653}  \\ \midrule    
    \edit{IP rand.}            & \edit{86.709}          & \edit{0.0753}    & \edit{0.0746}     & \edit{-0.658}    & \edit{-0.658}  \\ \midrule
    \edit{SA rand.}            & \edit{67.730}          & \edit{0.0676}    & \edit{0.0672}     & \edit{-0.749}    & \edit{-0.752}  \\ \midrule
    \edit{BO rand.}            & \edit{74.581}          & \edit{0.0711}    & \edit{0.0705}     & \edit{-0.693}    & \edit{-0.696}  \\ \midrule

    RIG-tree       & 69.325          & 0.0690    & 0.0689     & -0.757    & -0.757  \\ \midrule
    Coverage       & 170.911         & 0.0981    & 0.0968     & -0.665    & -0.667  \\ \midrule
    \edit{Spiral}  & \edit{54.251}   & \edit{0.0595}    & \edit{0.0595}     & \edit{-0.793}    & \edit{-0.791}  \\ \midrule \bottomrule
\end{tabular}
}
\caption{Mean information metrics for all optimization methods, averaged over $30$ continuous mapping trials.
\edit{The suffixes `lat.' and `rand.' indicate initializations using the lattice and random approaches, respectively.}
The lowest uncertainties and errors obtained with the CMA-ES justify our proposed global optimization strategy.} \label{T:methods}
\end{table}
%

Comparing the lattice approach with the CMA-ES and \ac{IP} methods \edit{using this informed initialization (`lat')}
confirms that optimization reduces both uncertainty and error.
With the lowest values,
the CMA-ES performs best on all indicators as it searches globally to escape local minima.
\edit{Whereas the CMA-ES variants with smaller and intermediate step sizes behave similarly in this problem,}
using much larger steps in `CMA-ES $(10,12)$' ($> 33\%$ of the workspace extent co-ordinate-wise)
results in worse performance,
as these lead to large random fluctuations during the evolutionary search,
which slows down convergence.
This reflects the importance of selecting suitable step sizes to cover the application domain,
as discussed by~\citet{Hitz2015} and~\citet{Hansen2006}.
Surprisingly,
applying \ac{BO} \edit{with lattice-based initialization} yields mean metrics poorer than those of the lattice \edit{itself}.
We suspect this to be due to its high exploratory behavior
causing erratic paths similar to those of `CMA-ES $(10,12)$'.
Despite attempting several commonly used acquisition functions,
we found \ac{BO} to be the most difficult one to tune
for the highly nonlinear problem domain.

\edit{The fact that the CMA-ES remains as the most successful optimizer
using random initialization (`rand')
suggests that it is most robust to varying starting conditions.
Nonetheless,
the results show that the lattice search still contributes significantly
to finding a good initial solution in practice,
which underlines the benefits of our two-step approach.
Though it performs well with the lattice,
the \ac{IP} method using random initialization scores almost as poorly
as the random benchmark alone
since it only conducts a local search to refine the solution.
This implies that it is very dependent on the initial grid selection
within our planner,
and further motivates global optimization routines.}

\subsubsection{Adaptive replanning evaluation} \label{SSS:adaptive_replanning}
\edit{We examine the performance of our online framework
in planning setups with adaptivity requirements (\Cref{SSS:adaptivity_requirements})
to assess its ability to \editrev{learn and} focus on targeted regions of interest
in different environments.
The experiments consider} two continuous mapping scenarios:
(a) `Split', handcrafted maps where the interesting area is well-defined
and (b) `Gaussian', the uniformly distributed fields from \Cref{SSS:benchmark_comparison}.
\editrev{Practically speaking,
`Gaussian' maps are representative of cases where the underlying field varies smoothly,
matching the assumption of the model,
whereas `Split' maps highlight
the targeted behaviour of the planner where specific zoning might be present in the field.
Both of these scenarios are relevant for an agricultural monitoring application as in \Cref{S:field_deployments}, for example,
depending on the type of plants growing on a field and the treatments it has received.}
`Split' maps are partitioned spatially such that half of the cells in $\mathcal{X}$
are classified as interesting based on \Cref{E:adaptive_planning}
with a base threshold of $\mu_{th} = 40\%$ and $\beta = 3$.
Then we apply these parameters for adaptive replanning
in the simulation setup from \Cref{SSS:benchmark_comparison}.

The gain of replanning online is evaluated by comparing a \edit{\emph{targeted} variant of our approach, i.e., using \Cref{E:adaptive_planning} as the planning objective,
against itself aiming for pure \emph{non-targeted} exploration, i.e., using \Cref{E:continuous_info_objective}, which treats} the information acquired from all locations in $\mathcal{X}$ equally.
As before, we perform $30$ \edit{simulation} trials in $30$\,m$\times$\,$30$\,m environments.
\begin{figure*}[!h]
\captionsetup[subfigure]{labelformat=empty}
\captionsetup[subfigure]{skip=-10pt}

 \begin{subfigure}{\columnwidth}
 \begin{subfigure}{0.48\columnwidth}
  \includegraphics[width=\columnwidth]{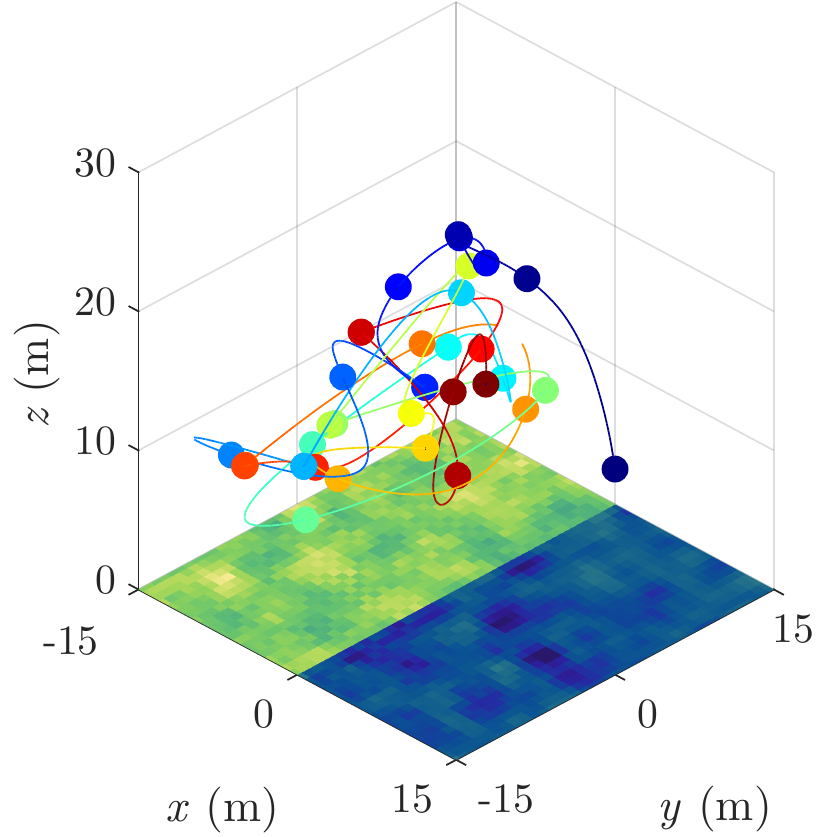}
 \end{subfigure}%
 \begin{subfigure}{0.48\columnwidth}
 \centering
  \includegraphics[width=\columnwidth]{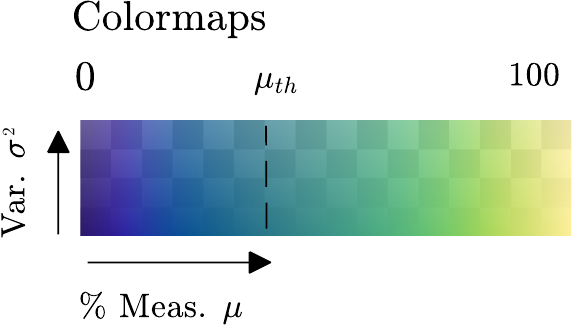} \\
 \vspace{1.5mm}
 \hspace{3mm}
  \includegraphics[width=0.85\columnwidth]{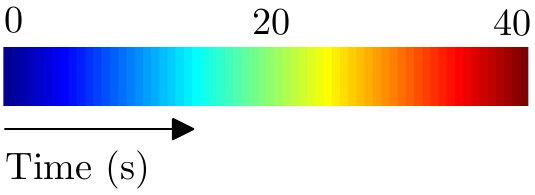}%
 \end{subfigure} \caption{} \label{SF:split_example}
 \end{subfigure}%
 \begin{subfigure}{0.96\columnwidth}
 \vspace*{7mm}%
 \hspace*{4mm}%
 \centering
 \begin{subfigure}{0.47\columnwidth}
  \includegraphics[width=\columnwidth]{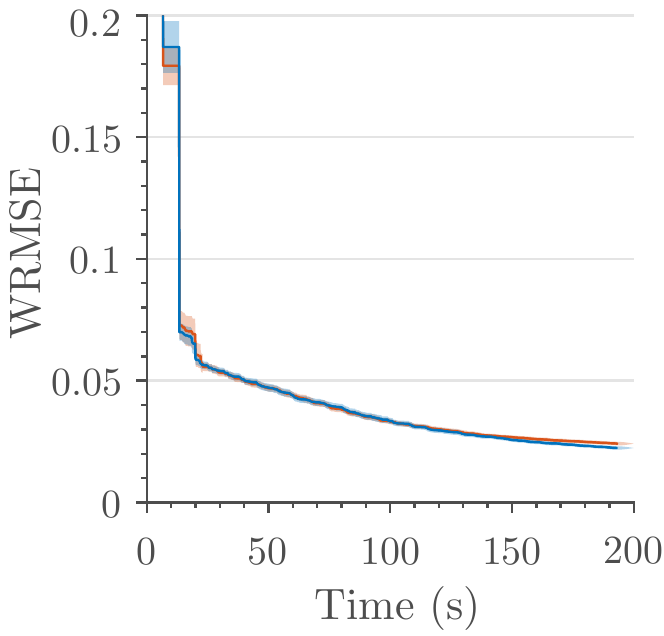}
 \end{subfigure}%
 \hspace*{1.9mm}
 \begin{subfigure}{0.48\columnwidth}
  \includegraphics[width=\columnwidth]{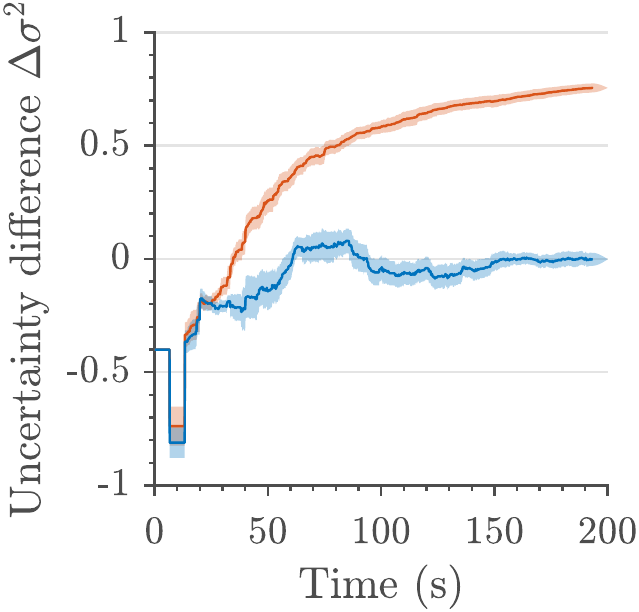}
 \end{subfigure} \caption{} \label{SF:adaptive_planning_evaluation_split}%
 \end{subfigure}
 \begin{minipage}{\columnwidth}
 \centering
 \hfill
 \end{minipage}%
 \begin{minipage}{\columnwidth}
 \centering
  (b)
 \end{minipage}%
 \vspace*{0.1mm}
 \hspace*{6mm}%
 \begin{minipage}{0.364\columnwidth}
  \vspace*{-8mm}
  \includegraphics[width=\columnwidth]{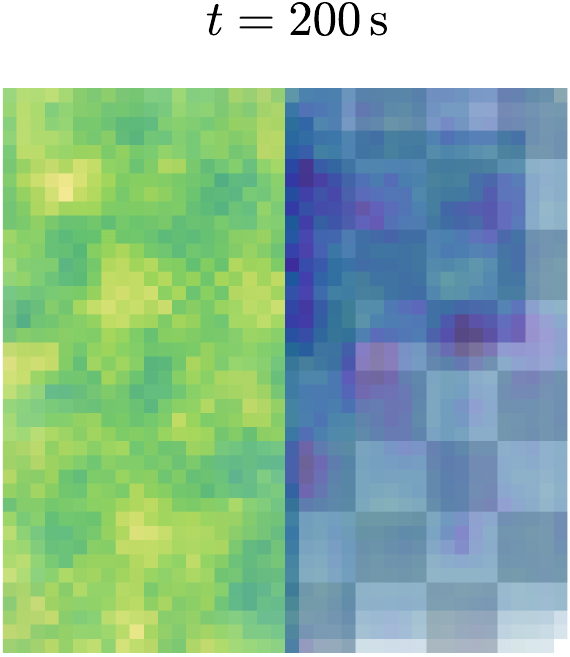}
 \end{minipage}%
 \hspace*{6mm}%
 \begin{minipage}{0.421\columnwidth}
 \vspace*{-5mm}%
  \includegraphics[width=\columnwidth]{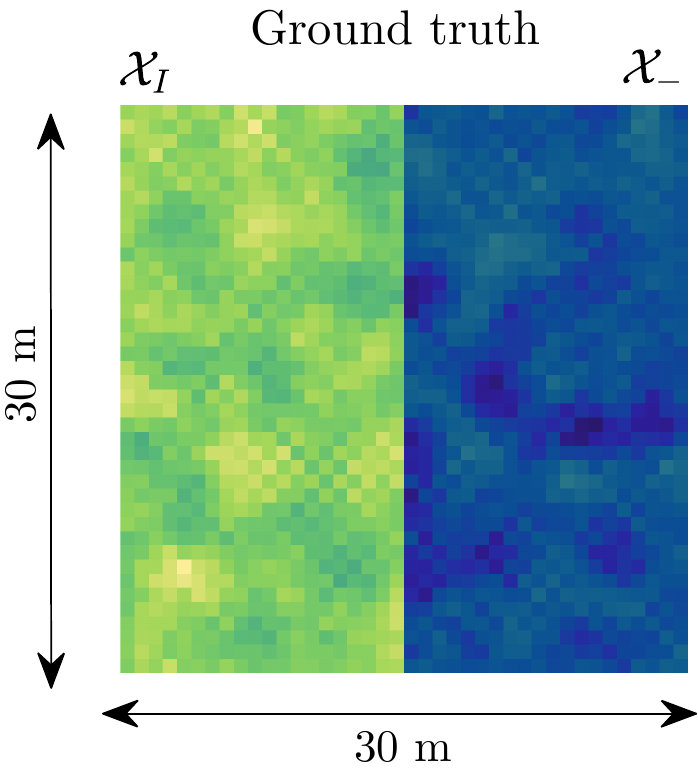}
 \end{minipage}%
 \hspace*{6mm}%
 \begin{subfigure}{0.96\columnwidth}
 \centering%
 \hspace*{4mm}%
 \begin{subfigure}{0.47\columnwidth}
  \centering
  \hspace*{8mm} Split
  \vspace*{2mm}

  \includegraphics[width=\columnwidth]{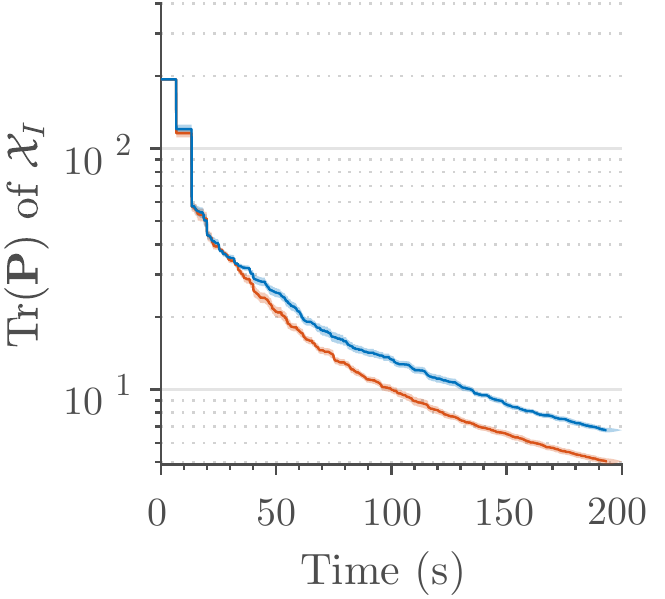}
 \end{subfigure}%
 \hspace*{1.9mm}
 \begin{subfigure}{0.47\columnwidth}
  \hspace*{16mm} Gaussian
  \vspace*{2mm}

  \includegraphics[width=\columnwidth]{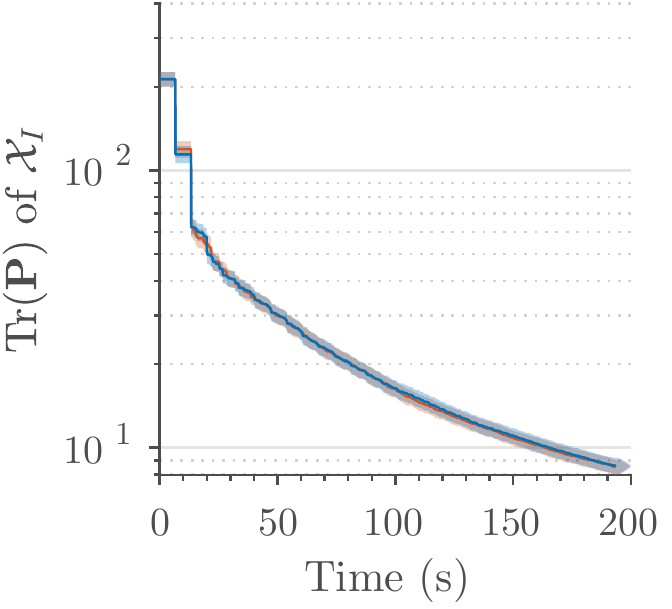}
 \end{subfigure}%
 \vspace{2.2mm}
 \hspace*{9mm}
 \begin{subfigure}{0.7\columnwidth}
  \includegraphics[width=\columnwidth]{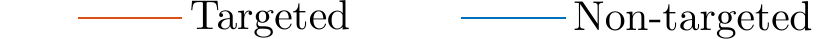}
 \end{subfigure} \caption{} \label{SF:adaptive_planning_evaluation}
 \end{subfigure}
 \vspace*{2mm}

 \begin{minipage}{\columnwidth}
 \centering
  (a)
 \end{minipage}%
 \begin{minipage}{\columnwidth}
 \centering
  (c)
 \end{minipage}%
 \vspace*{1.5mm}
 \hfill
 \caption{Evaluation of our adaptive replanning scheme.
 (a) \emph{Top}: Example `Split' scenario visualizing the trajectory traveled by our planner \edit{using the targeted objective in \Cref{E:adaptive_planning}} (colored line) in a $200$\,s mission.
 The spheres indicate measurement sites and the ground truth map is rendered.
 Colormaps are on the right.
 The dashed line shows the threshold $\mu_{th} = 40\%$ above which map regions are considered interesting (yellower).
 \emph{Bottom}: Final map output by our planner (left) compared with ground truth (right).
 The opacity indicates model uncertainty 
 with the checkerboard added for visual clarity.
 Lower opacity confirms higher certainty in the interesting area $\mathcal{X}_I$.
 (b) In `Split' scenarios,
 adaptivity achieves low error (left) with higher uncertainty differences (right) in interesting areas.
 (c) In `Split' scenarios (left),
 adaptivity reduces uncertainty faster in interesting areas,
 while performing comparably to a standard \edit{exploration} approach in `Gaussian' scenarios (right).
 In (b) and (c),
 the solid lines represent means over $30$ trials. The shaded regions depict $95\%$ confidence bounds.
 } \label{F:adaptive_planning}
\end{figure*}

For a quantitative evaluation,
we consider the variations of WRMSE and the uncertainty difference $\Updelta\sigma^2$
in the area of interest and the rest of the total area,
which is defined by~\citep{Hitz2015} as:
\begin{equation}
 \Updelta\sigma^2 = \frac{\bar{\sigma}^2(\mathcal{X}_{-}) -
     \bar{\sigma}^2(\mathcal{X}_{I})}{\bar{\sigma}^2(\mathcal{X}_{-})} \, \textrm{,}
\end{equation}
where $\bar{\sigma}^2($\textperiodcentered$)$ evaluates the mean variance
and $\mathcal{X}_{-}$ and $\mathcal{X}_{I}$ denote the sets of uninteresting and interesting locations, respectively,
\editrev{as described in \Cref{SSS:adaptivity_requirements}}.

Moreover,
the rate of uncertainty ($\Tr(\mathbf{P})$) reduction in $\mathcal{X}_{I}$
evaluates the ability of the planners to focus on interesting regions.

Our results are summarized in \Cref{F:adaptive_planning}.
As shown qualitatively in \Cref{SF:split_example},
once the uninteresting (bluer) map side $\mathcal{X}_-$ is classified in a `Split' environment,
planning adaptively \edit{in a targeted manner} leads to more measurements on the interesting (yellower) side $\mathcal{X}_I$,
which induces lower final uncertainty in this area, as expected.
\Cref{SF:adaptive_planning_evaluation_split} confirms that, in these scenarios,
the relative uncertainty difference $\Updelta\sigma^2$ increases more rapidly using adaptivity,
while map WRMSE remains low.
Note that,
early in the mission ($ < 30$\,s), both approaches behave similarly
to explore the initially unknown map.
Finally, \Cref{SF:adaptive_planning_evaluation} shows that
the benefit of adaptivity, in terms of reducing uncertainty in areas of interest,
is higher in `Split' environments when compared with `Gaussian'.
Since the region $\mathcal{X}_I$ is clearly distinguished,
purely informative measurements can be taken within the camera \ac{FoV}
given the thresholded objective.
Planning adaptively, however,
yields no disadvantages
when the field is uniformly dispersed.

\subsection{RIT-18 mapping scenario} \label{SS:dataset_evaluation}

\begin{figure}[h]
  \centering
  \begin{minipage}{0.47\columnwidth}
  \begin{subfigure}[]{\columnwidth}
   \includegraphics[width=\columnwidth]{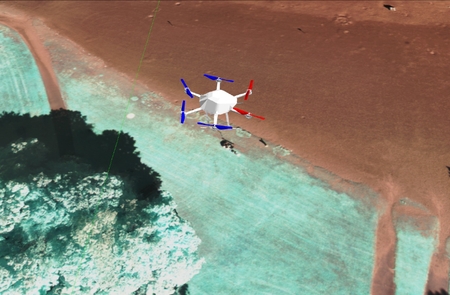}
  \caption{} \label{SF:gazebo_uav}
  \end{subfigure}
  \begin{subfigure}[]{\columnwidth}
   \includegraphics[width=\columnwidth]{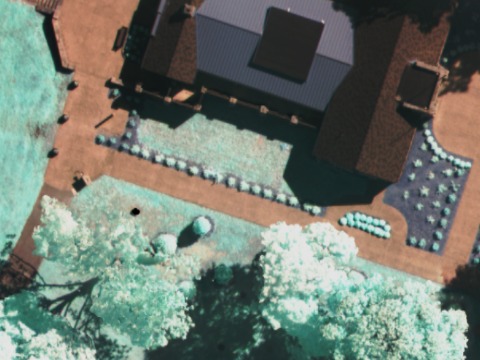}
  \caption{} \label{SF:camera_view}
  \end{subfigure}
  \end{minipage}
  \begin{subfigure}[]{0.502\columnwidth}
   \includegraphics[width=\columnwidth]{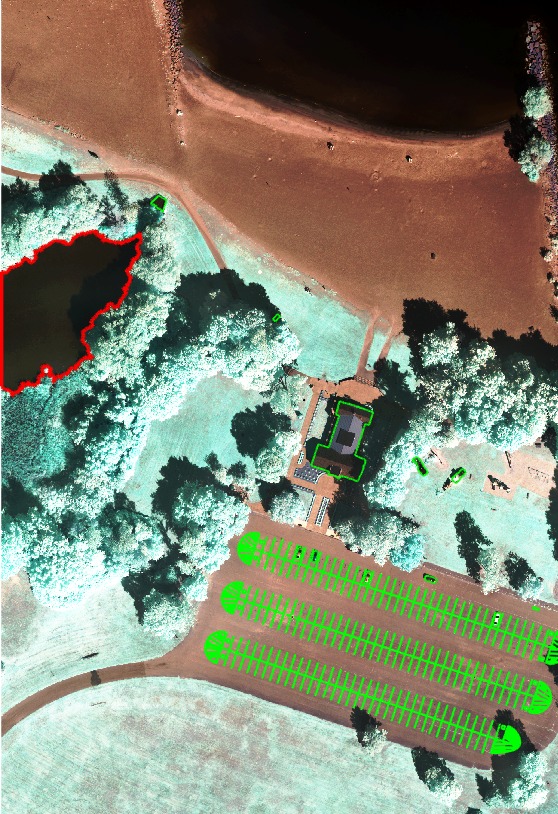}
  \caption{} \label{SF:rit18_ortho}
  \end{subfigure}
   \caption{Our photorealistic simulation setup in RotorS.
   (a) and (b) depict the AscTec Firefly \ac{UAV}
   and the view from its on-board camera.
   (c) shows an aerial view of the $200$\,m$\times$\,$290$\,m surveyed area (RIT-18 validation orthomosaic).
   The red and green lines annotate the two target classes for mapping using our approach:
   `Lake' and `BRV', respectively.}\label{F:gazebo_setup}
\end{figure}

We demonstrate our complete framework on a photorealistic scenario
in the Gazebo-based RotorS simulation environment~\citep{Furrer2016}.
In contrast to the preceding section,
these experiments show our framework in the discrete mapping domain
to reflect the nature of the target dataset.
\Cref{F:gazebo_setup} depicts our experimental setup,
which runs on a single desktop with a $2.6$\,GHz Intel i7 processor and $16$\,GB of RAM.
The planning and mapping algorithms were implemented in MATLAB on Ubuntu Linux
and interfaced to the Robot Operating System.
For mapping, we use RIT-18~\citep{Kemker2018},
a high-resolution 6-band VNIR dataset for semantic segmentation
consisting of coastal imagery along Lake Ontario in Hamlin, NY.
In our simulations,
the surveyed region is a $200$\,m\,$\times$\,$290$\,m area
featuring the RIT-18 validation fold.

\begin{figure}[h]
  \centering
  \begin{subfigure}[]{0.49\columnwidth}
   \includegraphics[width=\columnwidth]{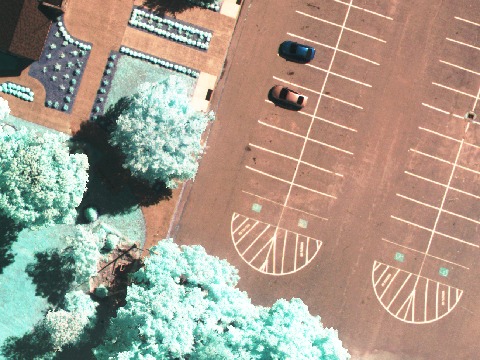}
  \caption{} \label{SF:image_rgb}
  \end{subfigure}
  \begin{subfigure}[]{0.49\columnwidth}
   \includegraphics[width=\columnwidth]{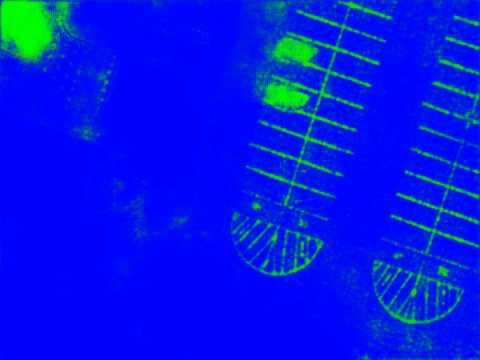}
  \caption{} \label{SF:image_out}
  \end{subfigure}
   \caption{Example classification result from an altitude of $70$\,m.
   (a) shows the RGB image channel input, and (b) visualizes the dense segmentation output.
   In (b), the probabilistic output for each class [`Lake', `BRV', `Bg']
   is mapped to the corresponding pixel intensity on the [R, G, B] channels.
   }\label{F:segnet_out}
\end{figure}

Our \ac{UAV} model is an AscTec Firefly equipped with a downward-facing camera,
which has a $360$\,px$\times$\,$480$\,px image resolution and a $(35.4\degree$,\,$47.2\degree)$ \ac{FoV}
in the $x$- and $y$-directions, respectively.
To extract measurements for active classification,
we use a modified version of the SegNet convolutional architecture~\citep{Badrinarayanan2017,Sa2018}
accepting multispectral as well as RGB image inputs.
The imagery registered from a given \ac{UAV} pose
is passed to the network to produce a dense semantic segmentation output,
as exemplified in \Cref{F:segnet_out}.
We simplify the classification problem
by only mapping the following 3 classes derived from the RIT-18 labels:
(a) `Lake'; (b) a combination of `Building', `Road Markings', and `Vehicle' (`BRV');
and (c) `Background' (`Bg'), i.e., all others.
These particular labels were chosen based on their distributions
to obtain strongly altitude-dependent classification performance,
as relevant for the \edit{terrain monitoring} problem setup.

\begin{figure}[h]
  \centering
  \begin{subfigure}[]{0.45\columnwidth}
   \includegraphics[width=\columnwidth]{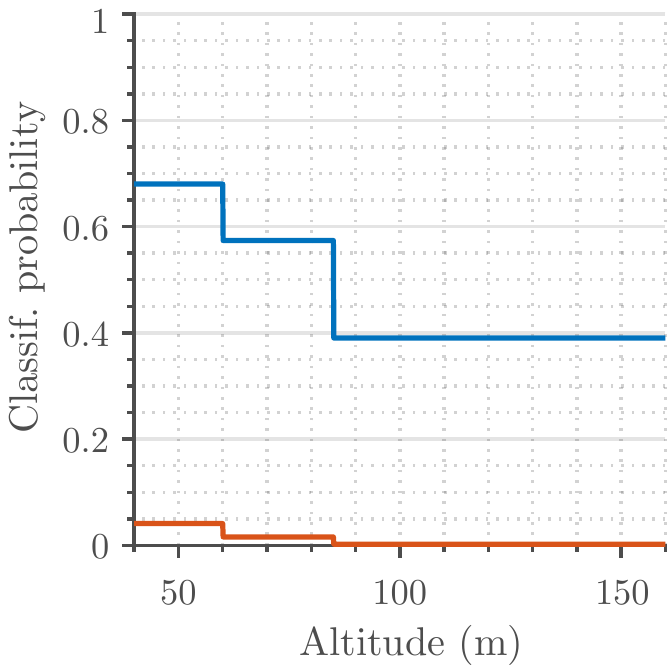}
  \caption{} \label{SF:sensor_class1}
  \end{subfigure}
  \begin{subfigure}[]{0.45\columnwidth}
   \includegraphics[width=\columnwidth]{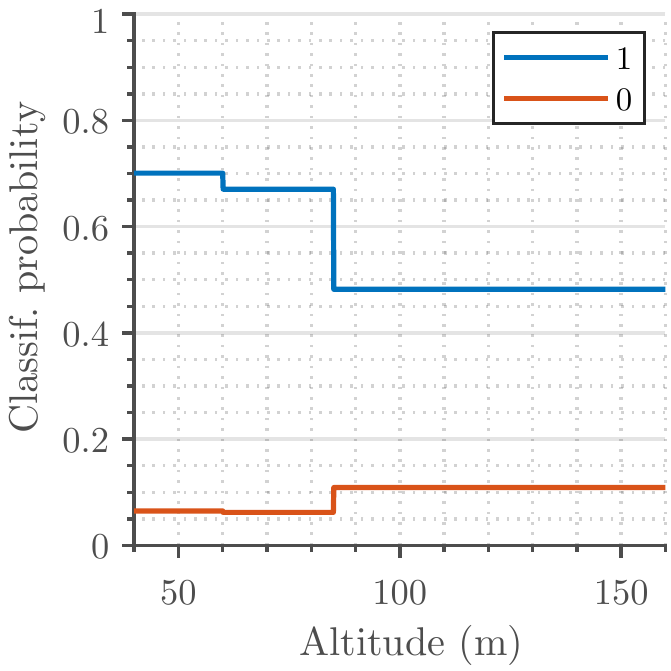}
  \caption{} \label{SF:sensor_class2}
  \end{subfigure}
   \caption{Sensor models for our trained classifier
   for the (a) `Lake' and (b) `BRV' classes.
   The blue and orange curves depict the probability of observing class label `1,
   given that the map contains `1 or `0', i.e., true and false positives, respectively.
   Note that the false positive probability can decrease with altitude as the
   classifier becomes more conservative with true outputs.}\label{F:sensor_segnet}
\end{figure}

To model the sensor for predictive planning,
we first trained SegNet on all labels using RIT-18 training fold imagery.
Our training procedure uses $(323,\,72,\,16)$ images
simulated at 3 different altitudes, $(50$\,m,\,$70$\,m,\,$100$\,m),
and is performed on an Nvidia Titan X Pascal GPU module.
At $70$\,m and $100$\,m,
the training images were additionally downscaled to exaggerate the effects of pixel mixing
at lower resolutions.
Then, classification accuracy was assessed
by using validation data to compute confusion matrices at each altitude for the 3 classes of interest
($30$\% train and $70$\% test split, with a higher proportion of training data at lower altitudes).
This enabled us to derive the sensor models in \Cref{F:sensor_segnet},
in which we associate intermediate altitudes with the closest performance statistics available.
Note that the altitude range considered here
is wider compared to the previous sections
due to the larger environment size.

We employ a discrete strategy to map the target region,
maintaining one independent occupancy grid layer for each of the 3 classes.
Each layer has a uniform resolution of $5$\,m,
and all cells are initialized with an uninformed probability of 0.5.
For predictive measurements when planning,
we use the sensor models in \Cref{F:sensor_segnet} conditioned on the most probable current map states.
For fusing new data,
we project the classifier output in \Cref{SF:image_out} on the occupancy grids for each class,
performing likelihood updates with the maximum pixel probabilities mapping to each cell.
Note that, unlike the pixel-wise classifier output,
our mapping strategy does not enforce the probabilities of a cell across the layers to sum to 1,
as a cell may contain objects from multiple classes.

The planning goal in this setup is to efficiently map the `BRV' class,
which would be useful, e.g., for identifying man-made features in search and rescue scenarios.
Our proposed approach with the CMA-ES is evaluated against
the ``lawnmower'' coverage strategy,
considered as the na\"ive choice of algorithm for such applications.
To investigate height-dependent performance,
trials are performed with two coverage patterns at fixed altitudes of $157$\,m and $104$\,m,
denoted `Cvge. 1' and `Cvge. 2', respectively.
In addition, we study both \edit{targeted} and \edit{non-targeted} versions of our approach,
in order to expose the benefits of using adaptive replanning to map  the regions of interest.
Our performance metrics are map entropy and RMSE with respect to the RIT-18 ground truth labels.

All methods are given a $400$\,s budget $B$.
To limit computational load on the classifier,
we assign a measurement frequency of $0.1$\,Hz,
allowing the \ac{UAV} to stop while processing images. 
As before,
trajectory optimization is performed on polynomials of order $k = 12$.
The \ac{UAV} starting position in our approach is set as $(33$\,m,\,$46$\,m) within the lower-left field corner with $104$\,m altitude
to achieve consistency with the lower-altitude coverage pattern.
For planning, we use polynomials defined by $N = 5$ waypoints
with a reference velocity and acceleration of $15$\,m$/$s and $20$\,m$/$s$^2$.
The 3-D grid search is executed on a scaled version of the 30-point lattice in \Cref{SF:lattice30},
stretched to cover the rectangular area,
and the CMA-ES optimizer runs with initial step sizes of $(50$\,m,\,$60$\,m,\,$40$\,m).
We apply a low threshold of $p_{th} =  0.4$ in \Cref{E:discrete_info_objective}
on the occupancy grid layer of the \edit{target `BRV' class
to define the adaptive planning requirement.}
The coverage benchmarks are designed based on the principles discussed in the preceding sections.

\begin{figure}[!h]
 \centering
  \includegraphics[width=\columnwidth]{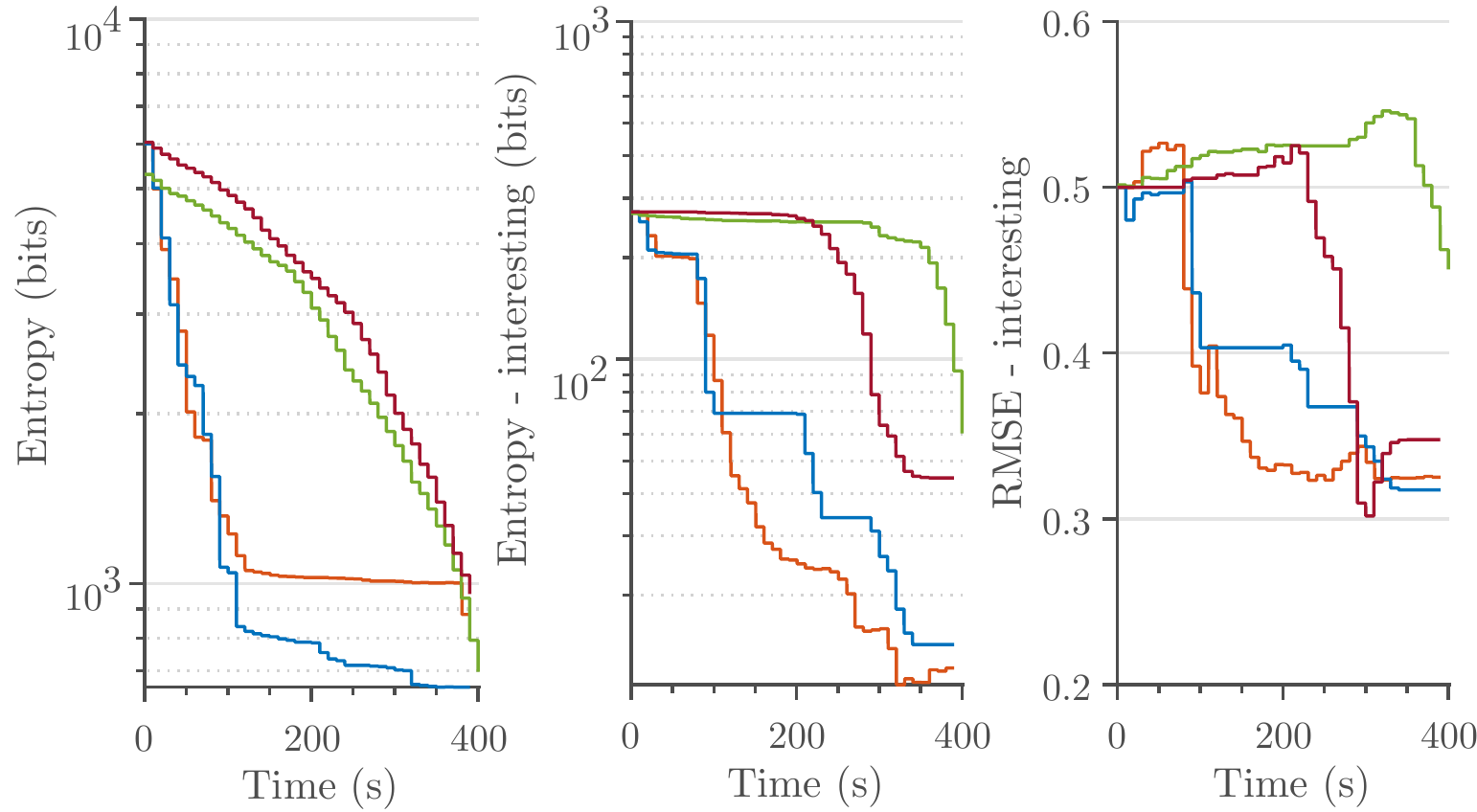}  \\
  \vspace{0.1cm}
  \includegraphics[width=0.88\columnwidth]{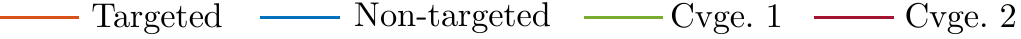}
  \caption{Comparison of our IPP approach using the CMA-ES against fixed-altitude coverage benchmarks (`Cvge. 1' = $157$\,m, `Cvge. 2' = $104$\,m)
  in a $400$\,s photorealistic mapping scenario.
  By planning adaptively,
  map uncertainty (middle) and error (right) in interesting areas (`BRV' class) reduce most rapidly,
  while yielding higher overall map uncertainty (left).
  Note the logarithmic scale of the Entropy axis.
  }\label{F:rit18_methods}
\end{figure}
\Cref{F:rit18_methods} compares the performance of each planner in this scenario.
As in \Cref{SSS:benchmark_comparison}, total map uncertainty reduces uniformly using the coverage patterns,
with interesting areas surveyed only towards the end due to the environment layout.
In these regions,
`Cvge. 2' (dark red) achieves higher-quality mapping than `Cvge. 1' (green)
as its lower altitude permits more accurate measurements.
This evidences the height-dependent nature of the classifier,
which motivates using \ac{IPP} to navigate in 3-D space.
By planning adaptively using our \edit{targeted} approach (orange),
both uncertainty and error decay most rapidly in the areas of interest,
as expected.
However, a \edit{non-targeted} strategy (blue) performs better in terms of overall map uncertainty,
since it is biased towards pure exploration.
These results demonstrate how our framework can be tailored to balance
exploration (uniform uncertainty reduction) and exploitation (mapping a target class)
in a particular scenario.
\begin{figure*}[!h]
\centering
\begin{subfigure}{.24\textwidth}
\includegraphics[width=\textwidth]{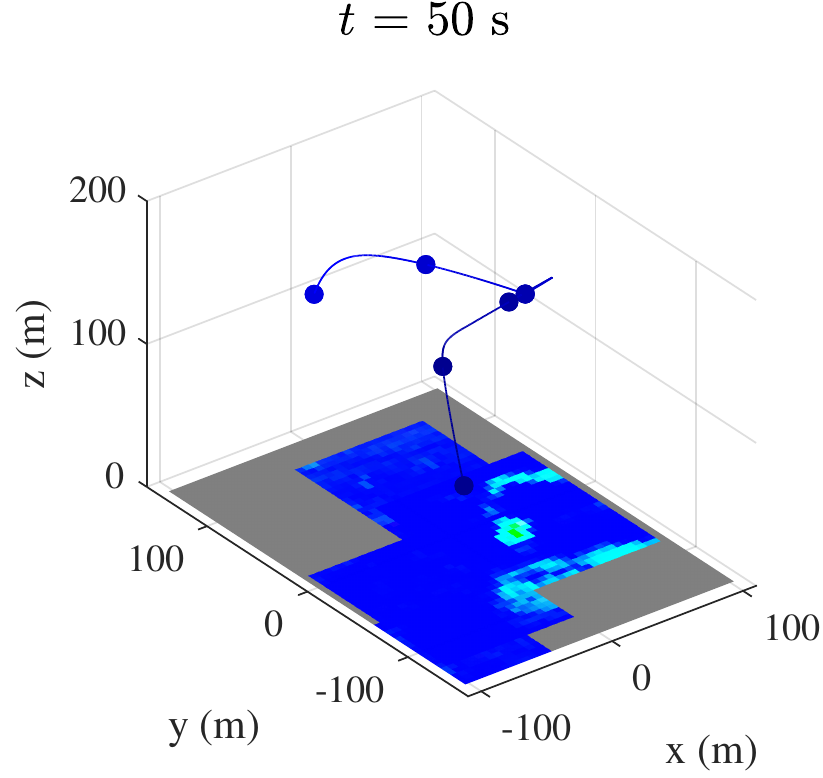}%
\end{subfigure} \hspace*{0.01mm}%
\begin{subfigure}{.24\textwidth}
\includegraphics[width=\textwidth]{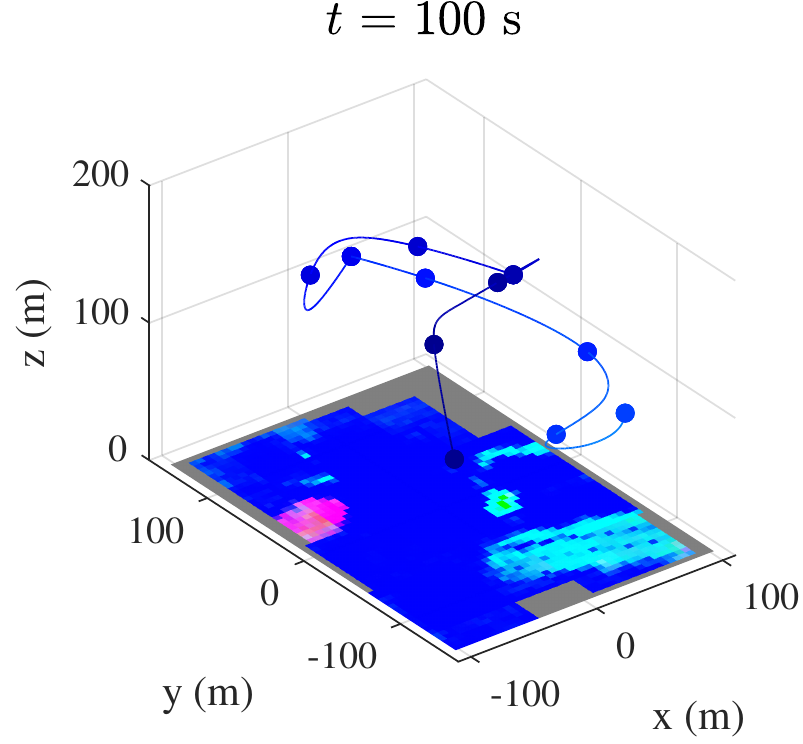}%
\end{subfigure} \hspace*{0.01mm}%
\begin{subfigure}{.24\textwidth}
\includegraphics[width=\textwidth]{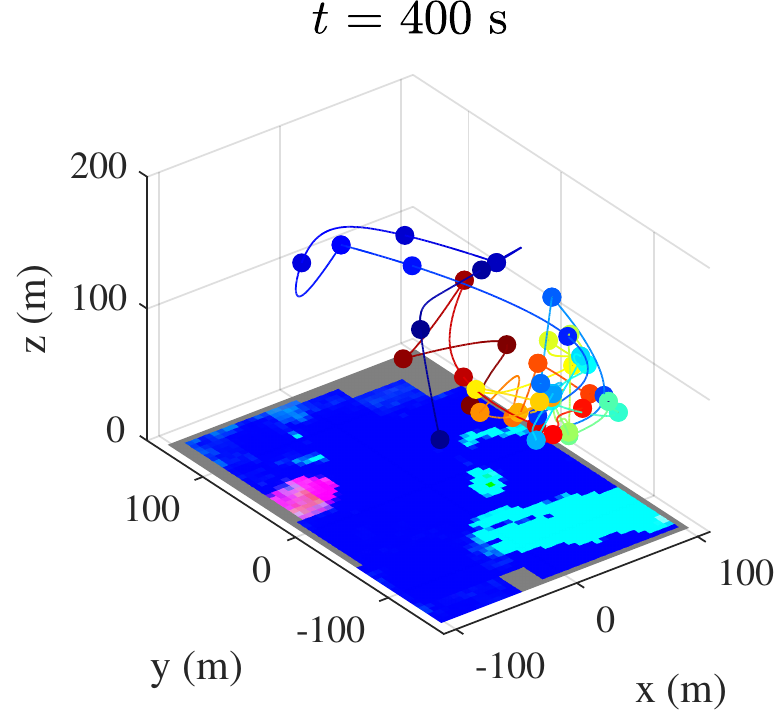}%
\end{subfigure} \hspace*{1mm}
\begin{subfigure}{.055\textwidth}
\includegraphics[width=\textwidth]{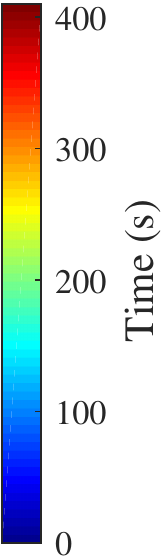}%
\end{subfigure}\hfill%
\begin{subfigure}{.16\textwidth}
\includegraphics[width=\textwidth]{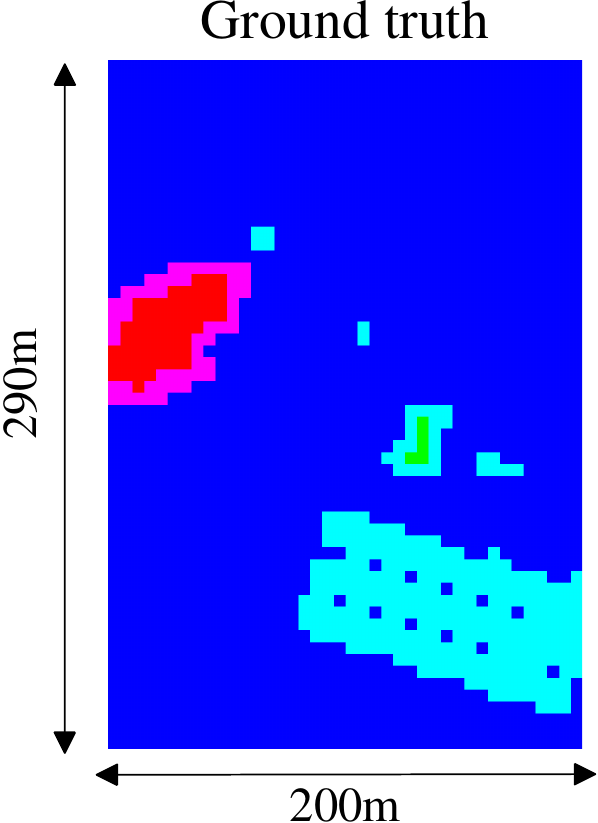}
\end{subfigure}%
\caption{Visualization of the trajectory traveled by our \edit{targeted} planner (colored line) in the $400$\,s mission.
The three plots depict different snapshots of the mission at times $t = 50$\,s, $100$\,s, and $400$\,s.
The RIT-18 ground truth is shown on the right.
In the trajectory plots,
the spheres indicate measurement sites and the current occupancy map states are rendered.
The colors portray composites of the three-layer map representation,
with the cell probabilities for each class [`Lake', `BRV', `Bg'] mapped to the corresponding intensities on the [R, G, B] channels.
Gray indicates unobserved space.
The sequence shows that our planner quickly explores the area to later focus on more closely mapping interesting regions (`BRV' class).
Note that the magenta and cyan cells indicate the presence of two classes, [`Bg', `Lake'] and [`Bg', `BRV'], respectively.
}
\label{F:rit18_ev}
\end{figure*}

\Cref{F:rit18_ev} visualizes the trajectory traveled by our \edit{targeted} planner during the mission.
As before,
the \ac{UAV} initially ($< 100$\,s) explores unobserved space,
before concentrating on high-probability areas for the `BRV' class once they have been discovered
(green, or cyan for cells containing both `BRV' and `Bg').
It can be seen that the map becomes more complete in these regions
as low-altitude measurements are accumulated.
Note that the two small cars to the right of the building and above the parking lot (visible in \Cref{SF:rit18_ortho})
are mapped incorrectly as our SegNet model is limited in segmenting out fine details
given the data it was trained on.
Considering the richness of the RIT-18 dataset,
an interesting direction for future work
is to explore different classification methods and target classes within our \ac{IPP} framework.

\edit{
\section{Field deployment} \label{S:field_deployments}
Finally,
we present experimental results from a field
deployment implementing our framework on a \ac{UAV} to monitor the vegetation distribution in a field.
The aim is \editrev{to} validate the system for performing a practical sensing task in a challenging outdoor environment,
with all algorithms running on-board and in real-time.

\begin{figure}[!h]
  \centering
  \begin{subfigure}[]{0.38\columnwidth}
  \includegraphics[width=\columnwidth]{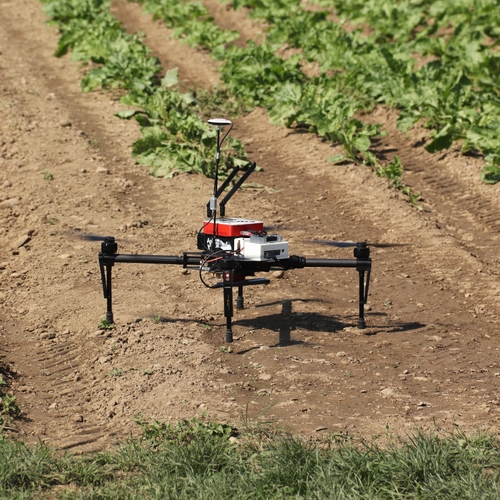}
  \caption{} \label{SF:flourish_uav_closeup}
  \end{subfigure}
  \begin{subfigure}[]{0.6\columnwidth}
  \includegraphics[width=\columnwidth]{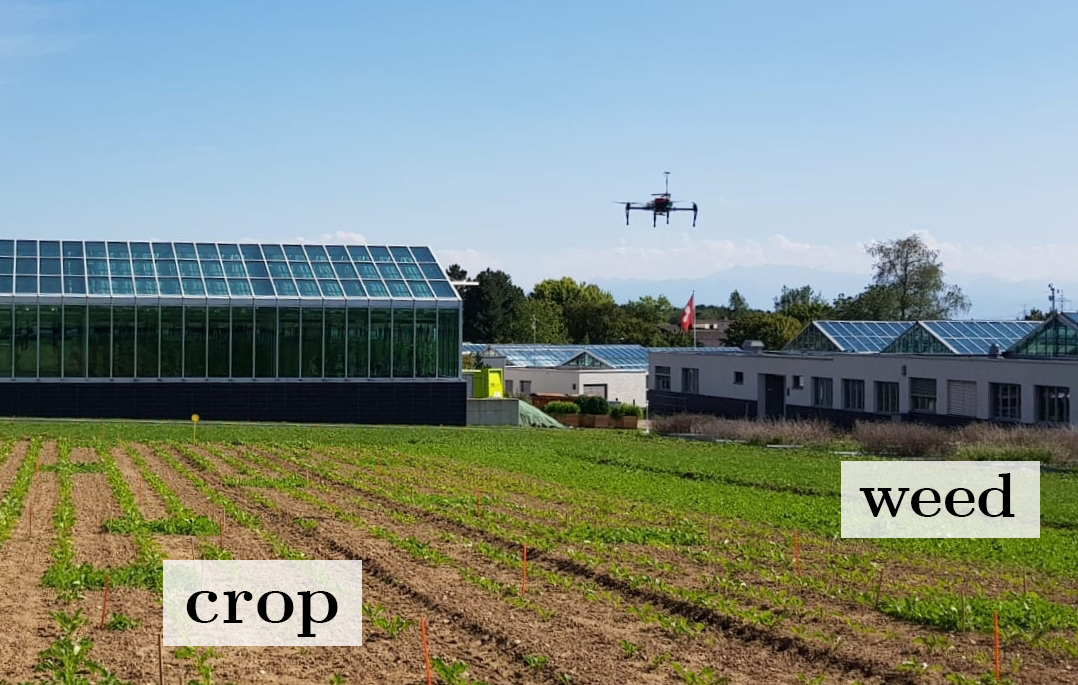}
  \caption{} \label{SF:flourish_uav_flying}
  \end{subfigure}
  
  \vspace{2mm}
  \begin{subfigure}[]{0.49\columnwidth}
  \includegraphics[width=\columnwidth]{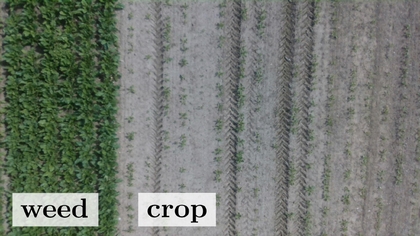}
  \caption{} \label{SF:flourish_camera_view1}
  \end{subfigure}
  \begin{subfigure}[]{0.49\columnwidth}
  \includegraphics[width=\columnwidth]{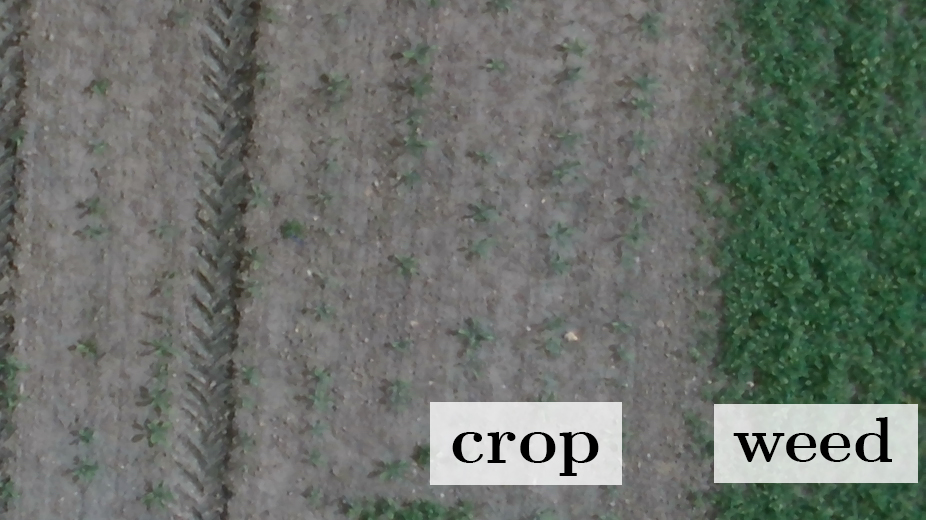}
  \caption{} \label{SF:flourish_camera_view2}
  \end{subfigure}
   \caption{\edit{(a) shows a close-up of the DJI Matrice M100 used in the outdoor trials.
   (b) illustrates the \ac{UAV} executing a mission, flying to monitor the vegetation distribution in a field of interest.
   (c) and (d) exemplify RGB images taken by the on-board camera at different altitudes, which are used for mapping online.
   Note that, in (b), the \ac{UAV} is positioned approximately above a boundary between the controlled crop (left) and weed (right) areas, visible in (c) and (d).}
   }\label{F:flourish_setup}
\end{figure}
The experiments were conducted on an agricultural field at the Research Station for Plant Sciences Lindau of ETH Zurich in Switzerland (Lat. $47.450040\degree$,\,Lon. $8.681056\degree$) in Sept. 2018. \Cref{F:flourish_setup} depicts our on-field setup. The \ac{UAV} platform in (a) is the DJI Matrice M100 monitoring a $20$\,m\,$\times$\,$20$\,m area within the field with maximum and minimum altitudes of $21$\,m and $8$\,m. As shown in (b), the controlled field features a central area of crops planted in row arrangements, surrounded by dense weed distributions on its edges. The width of the crop row area is $\sim 18$\,m. Vegetation mapping is performed using RGB imagery from a downward-facing Intel RealSenseZR300 camera with a resolution of $1920$\,px\,$\times$\,$1080$\,px and a \ac{FoV} of $(68.0\degree,\,47.2\degree)$. Example images taken from different altitudes are shown in (c) and (d). In these images, the boundaries between the weed and crop areas of the field can be distinguished.

We use the \ac{ROVIO} framework~\citep{Bloesch2015a} for state estimation
with \ac{MPC}~\citep{Kamel2017} to track trajectories output by the planner.
All computations, including modules for environmental mapping and informative planning, are based on our open-source package
and run on an on-board computer with a $3.2$\,GHz Intel NUC i7, $16$\,GB of RAM,
and running Ubuntu Linux 16.04 LTS with \ac{ROS} as middleware.
Further platform specifics are discussed by~\citep{Sa2017,Sa2018a}.

The goal is to map the level of \ac{ExG} in the area of interest based on the RGB images
and using our \ac{GP}-based mapping method for continuous variables.
It is defined by $ExG = 2g\,-\,r\,-\,b$,
where $r$, $g$, and $b$ are the normalized red, green, and blue color channels in the RGB space~\citep{Yang2015}.
Note that we normalized the range of the \ac{ExG}
based on the maximum and minimum magnitudes measured on the experimental field
in previously acquired datasets.

\begin{figure*}[!h]
  \begin{subfigure}{0.63\textwidth}
  \includegraphics[width=\textwidth]{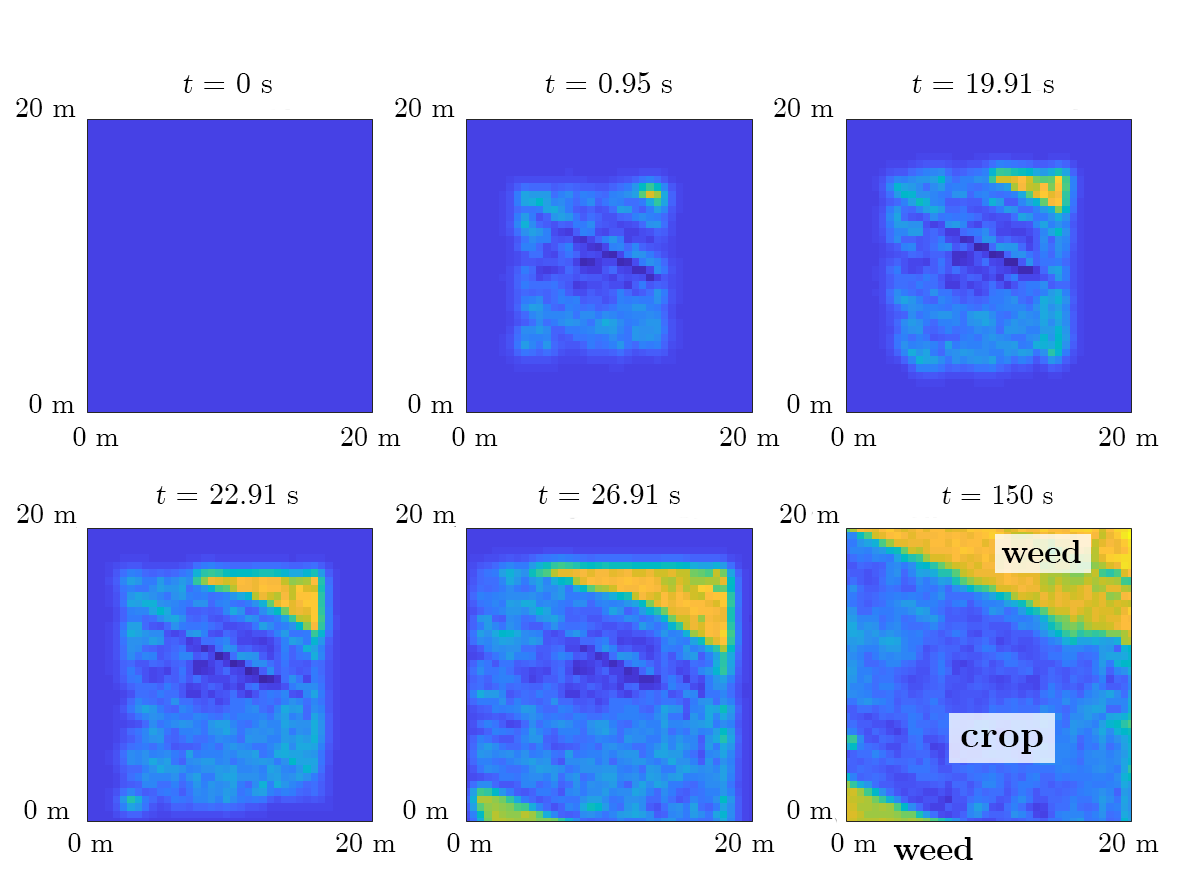}
  \end{subfigure}
  \begin{subfigure}{0.35\textwidth}
  \centering
  \vspace*{22mm}
  \hspace*{2mm}
  \includegraphics[width=0.75\textwidth]{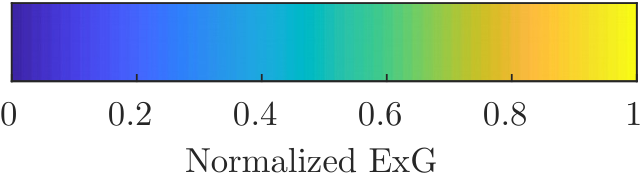} \\
  \vspace*{2mm}
  \includegraphics[width=\textwidth]{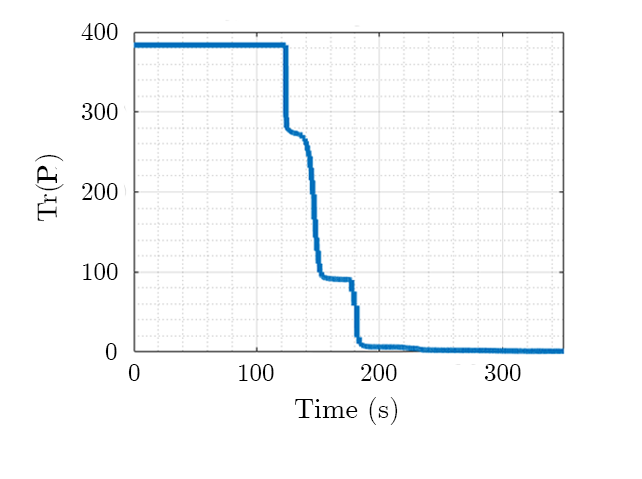}
  \end{subfigure}
   \caption{\edit{Experimental results from the field trial using our \ac{IPP} framework for \ac{UAV}-based vegetation monitoring. \emph{Top-right}: Colorbar, indicating the normalized Excess Green Index (ExG) level (plant greenness). \emph{Left}: Estimated map reconstructions (\ac{GP} means) of the normalized ExG on the field at different times $t$ during the mission. The bluer and yellower regions correspond to successfully identified crop row and weed regions, respectively. \emph{Bottom-right}: Evolution of total map uncertainty over time. The results confirm improving map completeness as the number of measurements increases.}} \label{F:flourish_result}
\end{figure*}
For mapping,
a uniform resolution of $0.5$\,m is set for both the training $\mathcal{X} $ and predictive $\mathcal{X}^*$ grids in the \ac{GP}.
Assuming that the field only contains soil,
we initialize the map with a uniform mean prior of $0$ normalized \ac{ExG}.
The isotropic Mat\'ern 3/2 kernel (\Cref{E:matern_kernel}) is applied
to approximate the vegetation spread in the field,
with the hyperparameters $\{\sigma_n^2,\,\sigma_f^2,\,l\} = \{0.50,\,0.50,\,1.76\}$ 
trained using images from a manually acquired dataset.
We recorded the dataset while flying the \ac{UAV} at a fixed altitude of $8$\,m over the area,
which corresponds to the minimum level allowed in the experiments,
and the maximum \ac{GSD} (in m$/$px) obtainable in the images,
i.e., highest possible mapping resolution in $\mathcal{X}$.

Measurements are extracted from RGB images taken at a constant frequency of $0.20$\,Hz.
We consider the sensor noise model defined by \Cref{E:sensor_continuous}
with coefficients $a=0.05$ and $b=0.2$ set to determine the variance variation
based on the altitude range.
Note that, due to the inherently high camera resolution, 
the scaling factor is $s_f=1$ across the altitude range.
To update the map with a new image,
the pixels are first projected on the ground plane 
based on the camera parameters and the estimated \ac{UAV} pose.
The projected pixels are averaged per cell,
and the normalized \ac{ExG} values for each cell are computed from the color channels.
Data fusion then proceeds according to the method described in \Cref{SSS:data_fusion}. 

The planning strategy considers
polynomial trajectories of order $k=12$ defined by $N=3$ waypoints
and optimized for a maximum reference velocity and acceleration of $5$\,m$/$s and $3$\,m$/$s$^2$.
The planning objective is set
as uncertainty reduction with no interest-based threshold (\Cref{E:continuous_info_objective}),
i.e., the aim is to reconstruct the field in a uniform manner, as quickly as possible.
We perform the 3-D grid search over a coarse 14-point lattice
and set initial CMA-ES step sizes of $4.5$\,m in each co-ordinate of the \ac{UAV} workspace.
For simplicity,
we do not specify a mission budget $B$;
instead allowing the algorithm to create fixed-horizon plans
until the mapping output is perceived visually as being complete.

The deployment results are reported in \Cref{F:flourish_result}\footnote{A video showing the \ac{UAV} trajectory is available at: \url{youtu.be/5dK8LcQH85o}}.
Since the ground truth data of normalized \ac{ExG} levels are not available,
we validate our framework by assessing the progression of total map uncertainty on the bottom-right,
which confirms that uncertainty is reduced over time.
Note that the curve in the plot is offset by $\sim 100$\,s as data recording was triggered
before the \ac{UAV} took off to take the first measurement.

Qualitatively,
the sequence of plots on the left verifies that
the estimated map does become more complete as images of the field are accumulated.
The yellower parts, corresponding to areas with high values of the normalized \ac{ExG},
indicate successfully identified weeds on the edges of the physical field (visible in \Cref{SF:flourish_camera_view1,SF:flourish_camera_view2}).
Towards the central area, a close look at the bluer parts with lower \ac{ExG} reveals that even the crop row details \textbf{}are mapped correctly.
These findings showcase our mapping strategy
and represent valuable data that could guide practical crop management decisions.
}

\section{Conclusion and future work} \label{S:conclusion}
This paper introduced a general \ac{IPP} framework for environmental monitoring applications using an aerial robot.
The method is capable of mapping either discrete or continuous target variables on a terrain
using variable-resolution data received from probabilistic sensors.
The resulting maps are employed for \ac{IPP}
by optimizing parameterized continuous-space trajectories initialized by a coarse 3-D search.

Our approach was evaluated extensively in simulations using synthetic and real world data.
The results reveal higher efficiency compared to state-of-the-art methods
and highlight its ability to efficiently build models with lower uncertainty in value-dependent regions of interest.
Furthermore, we validated our framework in an active classification problem
using a publicly available dataset.
These experiments demonstrated its online application on a photorealistic mapping scenario
with a SegNet-based sensor for data acquisition.
\edit{Finally, a proof of concept was presented
showing the algorithms running on-board and in real-time
for an agricultural monitoring task
in an outdoor environment.}

The implementation of the proposed planner is released for use 
and further development by the community
along with sample \edit{experimental} results.
Future theoretical work will investigate scaling the approach to larger environments
and extending the mapping model to capture temporal dynamics.
This would enable
previously acquired data to be used as a prior in persistent monitoring missions.
Towards more accurate map building in practice,
it would be interesting 
to also incorporate the robot localization uncertainty in the decision-making algorithm.

\begin{acknowledgements}
We would like to thank Dr. Frank Liebisch for his useful discussions.
This project has received funding from the European Union’s Horizon 2020 
research and innovation programme under grant agreement No 644227 and from the 
Swiss State Secretariat for Education, Research and Innovation (SERI) under 
contract number 15.0029.
\end{acknowledgements}

\bibliographystyle{apalike}
\bibliography{references/2017-tbd-popovic}

\end{document}

%% file: style/acronyms.tex
\begin{acronym}

\acro{2D}{two-dimensional}

\acro{3D}{three-dimensional}

\acro{ExG}{Excess Green Index}

\acro{ROVIO}{Robust Visual Inertial Odometry}

\acro{MPC}{Model Predictive Control}

\acro{SDK}{Software Development Kit}

\acro{AHRS}{attitude and heading reference system}

\acro{AUV}{autonomous underwater vehicle}

\acro{CPP}{Chinese Postman Problem}

\acro{DoF}{degree of freedom}
\acrodefplural{DoF}[DoFs]{degrees of freedom}

\acro{DVL}{Doppler velocity log}

\acro{FSM}{finite state machine}

\acro{IMU}{inertial measurement unit}

\acro{LBL}{Long Baseline}

\acro{MCM}{mine countermeasures}

\acro{MDP}{Markov Decision Process}
\acrodefplural{MDP}[MDPs]{Markov Decision Processes}

\acro{POMDP}{Partially Observable Markov Decision Process}
\acrodefplural{POMDP}[POMDPs]{Partially Observable Markov Decision Processes}

\acro{PRM}{Probabilistic Roadmap}
\acrodefplural{PRM}[PRM]{Probabilistic Roadmaps}

\acro{ROI}{region of interest}
\acrodefplural{ROI}[ROIs]{regions of interest}

\acro{ROS}{Robot Operating System}

\acro{ROV}{remotely operated vehicle}

\acro{RRT}{Rapidly-exploring Random Tree}
\acrodefplural{RRT}[RRTs]{Rapidly-exploring Random Trees}

\acro{SA}{Simulated Annealing}

\acro{SLAM}{simultaneous localization and mapping}

\acro{SSE}{sum of squared errors}

\acro{STOMP}{Stochastic Trajectory Optimization for Motion Planning}

\acro{TRN}{Terrain-Relative Navigation}

\acro{UAV}{Unmanned Aerial Vehicle}

\acro{USBL}{Ultra-Short Baseline}

\acro{IPP}{Informative Path Planning}

\acro{FoV}{Field of View}
\acrodefplural{FoV}[FoVs]{Fields of View}

\acro{CDF}{cumulative distribution function}

\acro{ML}{maximum likelihood}

\acro{RMSE}{Root Mean Squared Error}
\acro{MLL}{Mean Log Loss}

\acro{GP}{Gaussian Process}
\acrodefplural{GP}[GPs]{Gaussian Processes}

\acro{KF}{Kalman Filter}

\acro{IP}{Interior Point}
\acro{BO}{Bayesian Optimization}

\acro{GSD}{Ground Sample Distance}

\end{acronym}

%% file: additionals/gt_tikz.tex
\begin{tikzpicture}[scale=.2,every node/.style={minimum size=1cm}]

    \begin{scope}[
    	yshift=0,every node/.append style={
    	    yslant=0.5,xslant=-1},yslant=0.5,xslant=-1
    	             ]

        \fill[white,fill opacity=.9] (0,0) rectangle (5,5);
        \draw[black,very thick] (0,0) rectangle (5,5);
        \draw[step=5mm, black] (0,0) grid (5,5);
    \end{scope}

    \begin{scope}[
    	yshift=0,every node/.append style={
    	    yslant=0.5,xslant=-1},yslant=0.5,xslant=-1
    	             ]
    \end{scope}

\end{tikzpicture}

%% file: additionals/low_altitude_tikz.tex
\begin{tikzpicture}[scale=.2,every node/.style={minimum size=1cm},baseline]

    \begin{scope}[
    	yshift=0,every node/.append style={
    	    yslant=0.5,xslant=-1},yslant=0.5,xslant=-1
    	             ]

        \fill[white,fill opacity=.9] (0,0) rectangle (5,5);
        \draw[black,very thick] (0,0) rectangle (5,5);
        \draw[step=5mm, black] (0,0) grid (5,5);
        \draw[step=5mm, red,ultra thick] (1.499,1.499) grid (3.5,3.5);
    \end{scope}

    \begin{scope}[
    	yshift=0,every node/.append style={
    	    yslant=0.5,xslant=-1},yslant=0.5,xslant=-1
    	             ]
        \coordinate[] (O) at (0,0,-15);
        \coordinate[] (P1) at (1.5,1.5,0);
        \coordinate[] (P2) at (3.5,1.5,0);
        \coordinate[] (P3) at (1.5,3.5,0);
        \coordinate[] (P4) at (3.5,3.5,0);
    
        \draw[color=black] (O)--(P1);
        \draw[color=black] (O)--(P2);
        \draw[color=black] (O)--(P3);
        \draw[color=black] (O)--(P4);
        \draw[fill=gray,opacity=0.05] (O)--(P1)--(P2);
        \draw[fill=gray,opacity=0.05] (O)--(P2)--(P4);
        \draw[fill=gray,opacity=0.05] (O)--(P3)--(P4);
        \draw[fill=gray,opacity=0.05] (O)--(P1)--(P3);
        \draw[fill=blue,opacity=0.1] (P1)--(P2)--(P4)--(P3);
    \end{scope}
    
\end{tikzpicture}

%% file: additionals/high_altitude_tikz.tex
\begin{tikzpicture}[scale=.2,every node/.style={minimum size=1cm},baseline]

    \begin{scope}[
    	yshift=0,every node/.append style={
    	    yslant=0.5,xslant=-1},yslant=0.5,xslant=-1
    	             ]

        \fill[white,fill opacity=.9] (0,0) rectangle (5,5);
        \draw[black,very thick] (0,0) rectangle (5,5);
        \draw[step=5mm, black] (0,0) grid (5,5);
        \draw[step=10mm, red, ultra thick] (1,1) grid (4,4);
    \end{scope}

    \begin{scope}[
    	yshift=0,every node/.append style={
    	    yslant=0.5,xslant=-1},yslant=0.5,xslant=-1
    	             ]
        \coordinate[] (O) at (0,0,-20);
        \coordinate[] (P1) at (1,1,0);
        \coordinate[] (P2) at (4,1,0);
        \coordinate[] (P3) at (1,4,0);
        \coordinate[] (P4) at (4,4,0);
    
        \draw[color=black] (O)--(P1);
        \draw[color=black] (O)--(P2);
        \draw[color=black] (O)--(P3);
        \draw[color=black] (O)--(P4);
        \draw[fill=gray,opacity=0.05] (O)--(P1)--(P2);
        \draw[fill=gray,opacity=0.05] (O)--(P2)--(P4);
        \draw[fill=gray,opacity=0.05] (O)--(P3)--(P4);
        \draw[fill=gray,opacity=0.05] (O)--(P1)--(P3);
        \draw[fill=blue,opacity=0.1] (P1)--(P2)--(P4)--(P3);
    \end{scope}

\end{tikzpicture}